\documentclass[letterpaper, 10 pt, conference]{ieeeconf}  %

\IEEEoverridecommandlockouts                              %

\usepackage{graphicx}
\usepackage{amsmath,amssymb} %
\usepackage{color}
\usepackage{times}
\usepackage{float}
\usepackage{epsfig}
\usepackage{multirow}
\usepackage{pifont}
\usepackage{subcaption}
\usepackage{cite} 
\usepackage{adjustbox}
\usepackage{threeparttable}
\usepackage{xspace}
\usepackage[dvipsnames, table]{xcolor}
\usepackage{booktabs}
\usepackage{amsfonts}
\usepackage{wasysym}
\usepackage{placeins}
\usepackage{bbm}
\usepackage{tabularx}
\usepackage[hidelinks]{hyperref}
\usepackage[normalem]{ulem}

\newcommand{\ourdataset}{\textsc{ATG4D}}

\newcommand{\nuscenes}{\textsc{nuScenes}}

\DeclareRobustCommand\onedot{\futurelet\@let@token\@onedot}
\def\@onedot{\ifx\@let@token.\else.\null\fi\xspace}

\DeclareMathOperator*{\argmin}{arg\,min}

\title{\LARGE \bf The Importance of Prior Knowledge in Precise Multimodal Prediction}

\author{Sergio Casas$^{*1, 2}$, Cole Gulino$^{*1}$, Simon Suo$^{*1, 2}$ and Raquel Urtasun$^{1, 2}$\\
 Uber Advanced Technologies Group$^1$, University of Toronto$^2$\\
 \{sergio.casas, cgulino, suo, urtasun\}@uber.com
\thanks{*Denotes equal contribution.}%
}

\begin{document}
\maketitle
\thispagestyle{empty}
\pagestyle{empty}

\begin{abstract}
Roads have well defined geometries, topologies, and traffic rules. While this has been widely exploited in motion planning methods to produce maneuvers that obey the law, little work has been devoted to utilize these priors in perception and motion forecasting methods. 
In this paper we propose to incorporate these structured priors as a loss function. In contrast to imposing hard constraints, this approach allows the model to handle non-compliant maneuvers when those happen in the real world. 
Safe motion planning is the end goal, and thus a probabilistic characterization of the possible future developments of the scene is key to choose the plan with the lowest expected cost.
Towards this goal, we design a framework that leverages REINFORCE 
to incorporate non-differentiable priors over sample trajectories from a probabilistic model, thus optimizing the whole distribution.
We demonstrate the effectiveness of our approach on real-world self-driving datasets containing complex road topologies and multi-agent interactions. Our motion forecasts not only exhibit better precision and map understanding, but most importantly result in safer motion plans taken by our self-driving vehicle. We emphasize that despite the importance of this evaluation, it has been often overlooked by previous perception and motion forecasting works.

\end{abstract}

\section{Introduction} \label{intro}

Self-driving vehicles (SDV) have the potential to make a large impact on our society, making transportation safer, cheaper and more efficient. A core component of every self-driving vehicle is its ability to perceive the world (including dynamic objects), and forecast how the future might unroll.
In  recent years there has been incredible progress in perception systems \cite{li2016vehicle, yang2018pixor, shi2019part}.  Many challenges still remain, however, in providing motion forecasts that are simultaneously diverse and precise \cite{rhinehart2018r2p2}. That is,  having the ability to cover all the modes of the data distribution while generating few highly unrealistic trajectories.

\begin{figure}[t]
    \includegraphics[width=\linewidth]{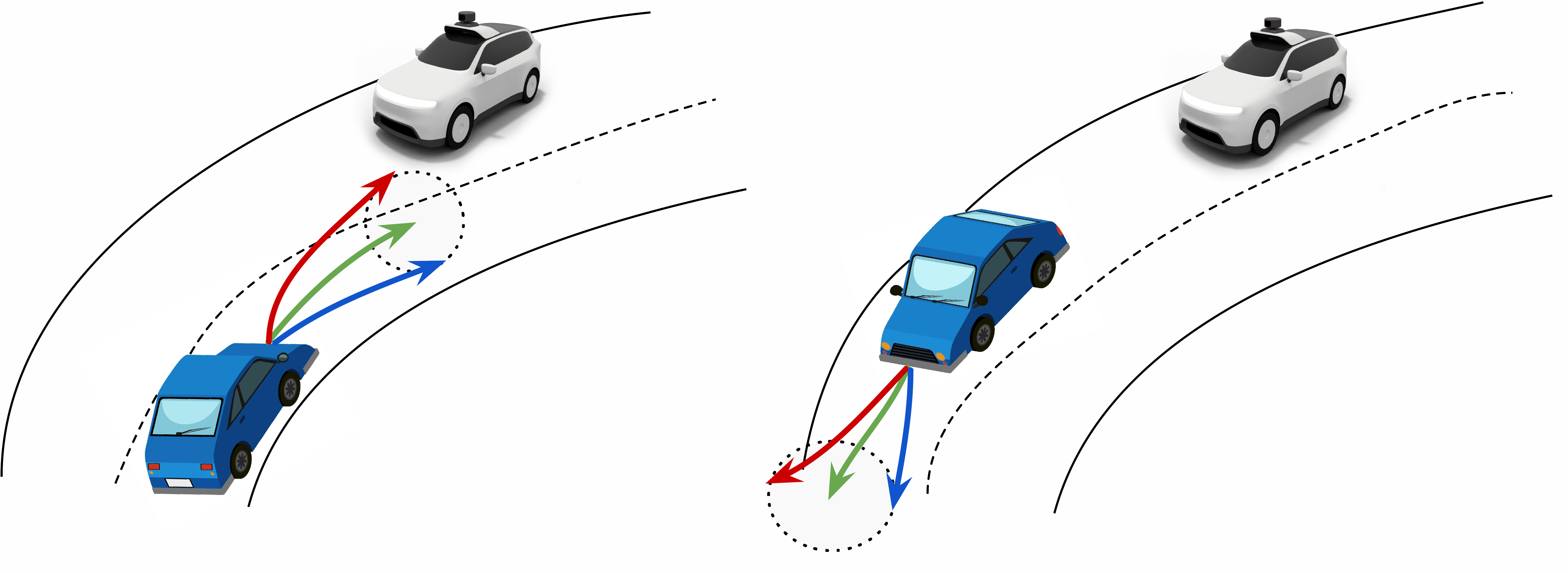}
    \caption{Importance of incorporating prior knowledge depicted in two scenarios. We show the \textbf{\textcolor{Green}{ground-truth trajectory}}, and two predictions with the same L2 error, one that is \textbf{\textcolor{BrickRed}{dangerous}} from the SDV perspective, and another one that is \textbf{\textcolor{NavyBlue}{harmless}}.
    }
    \label{primer}
    \vspace{-0.3cm}
\end{figure}

Roads in modern cities have well defined geometries, topologies, as well as traffic rules. 
The vast majority of actors in the scene will adhere to this structure, for example driving close to the middle of their lane, respecting stop signs, or obeying yielding laws. 
These agents will also most likely act in a socially acceptable manner, avoiding collisions with other traffic participants.
Despite this fact, most  perception and motion forecasting systems are trained to be as close as possible to the ground truth employing symmetric loss functions that do not take this structure into account. For example, euclidean distance at the waypoint level between the predicted and ground-truth future trajectories is a common choice for motion forecasting.  
This can cause uncomfortable rides for the self-driving vehicle with plenty of sudden brakes and steering changes due to false positive motion forecasts intruding into the ego-car's lane, such as the red trajectory illustrated in Fig.~\ref{primer}-Left. Even worse, these sudden reactions to avoid an imminent collision with a motion forecast can cause an actual collision (with the ground-truth) as a by-product. It is also critical to recall all actors (and their future motion) in the SDV lane, since otherwise it would look like free space ahead and the SDV could dangerously accelerate, as depicted by the red motion forecast in Fig.~\ref{primer}-Right.

One possible solution is to add the aforementioned intuitions into the motion forecasting model as hard constraints. Unfortunately, this may not be resilient to non-compliant behavior from other actors as well as map failures, predicting possibly unrealistic and dangerous situations.
In this paper we take an alternative approach and design loss functions that encourage our perception and prediction system to only violate these constraints when they happen in reality. 

Incorporating prior knowledge via loss functions is easy when the perception and prediction modules are deterministic.
However, deterministic systems fail to capture the inherent uncertainty of the future. This can be catastrophic when it fails to capture the true actor intention (e.g., crossing the street vs waiting, yielding vs not).  
In order to plan a safe maneuver, coverage of the possible future scenarios is required, along with information about the likelihood of each possible future such that the motion planner can choose the plan with the lowest expected cost. 
The Gaussian distribution and mixtures thereof have been widely used to represent uncertainty over spatial locations \cite{casas2019spatially, cui2019multimodal, chai2019multipath}. However, as shown in \cite{rhinehart2018r2p2}, maximizing the log likelihood of data encourages the model to produce distributions with high recall in order to avoid the big penalty associated with low-density areas. As a consequence  many unrealistic samples are generated, sacrificing the precision of the model. 

In this paper, we show that making explicit use of our prior knowledge about the geometry and topology of the roads as well as the traffic rules can provide more precise distributions over future outcomes while preserving their recall. However, this is challenging as these priors are typically non-differentiable and thus not directly amenable to gradient based optimization. For instance, the fact that humans tend to follow traffic rules can be better described as a discrete (follow / not follow) action.
To this end, we propose a flexible framework to incorporate non-differentiable prior knowledge as a loss and exploit the popular REINFORCE \cite{williams1992simple} gradient estimator. Our formulation allows us to optimize for any prior knowledge on future trajectories, as long as drawing samples from the perception and prediction model, and obtaining their likelihood can be done efficiently. 
In particular, we apply our formulation to model how the vehicles interact with the map, encouraging the predictions to respect lane dividers and traffic lights. We also exploit our framework to make the motion forecasting module more planning aware by emphasizing the importance of high recall and high precision near the SDV route.

Our experiments show that our proposed framework can improve the map understanding of state-of-the-art motion forecasting methods in very complex, partially observable urban environments in two challenging real-world datasets: ATG4D \cite{yang2018pixor} and nuScenes \cite{caesar2019nuscenes}. 
Importantly, our approach achieves significant improvements in the precision of the trajectory distribution, while maintaining the recall. 
Unlike previous works, in this paper we advocate for measuring the system-level impact of the motion forecasts via motion planning metrics. We demonstrate that including prior knowledge not only results in more comfortable rides, but also in major safety improvements over the decisions taken by a state-of-the-art motion planner \cite{sadat2019jointly}. Moreover, we show that achieving a lower minADE alone, the most frequently used metric to benchmark multi-modal motion forecasts, may not translate into safer motion plans.

\section{Related Work}

In order to plan a safe maneuver, we must effectively deal with noisy and partial observations from sensor data, and provide an accurate characterization of the distribution over future trajectories of all actors in the scene. We thus want to model  $p(Y|X)$, with $X=\begin{Bmatrix}x_1, x_2, \cdots, x_N\end{Bmatrix}$ the observations (i.e. a local context for each actor) and $Y = \begin{Bmatrix} y_1, y_2, \cdots, y_N \end{Bmatrix}$ the future trajectories of all $N$ actors.
This is particularly challenging since the future is inherently uncertain, and actors' discrete decisions induce highly multimodal distributions (e.g., turning right vs. going straight at an intersection, going vs. stopping at a yield sign).

In traditional self-driving stacks, there is an object detection module responsible for recognizing traffic participants in the scene, followed by a motion forecasting module that predicts how the scene might unroll, given the current state of each actor. However, the actor state is typically a very compact representation including pose, velocity, and acceleration. As a consequence, it is hard to incorporate uncertainty coming from sources such as sensor noise and occlusion.

FAF \cite{luo2018fast} unified these two tasks by having a single, fully convolutional backbone network predict both the current and future states for each pixel in a bird's eye view grid, directly from a voxelized LiDAR pointcloud. This naturally propagates uncertainty between the two tasks in the feature space, without any need for handcrafted, intermediate representations. \cite{casas2018intentnet} extended this framework to include the map as an input by adding a parallel fully convolutional backbone network to process a semantic raster map of the scene, thus making fusion trivial by concatenation. Recently, this framework was further extended to model agent-agent interactions via graph neural networks \cite{casas2019spatially}, learn a cost map for motion planning \cite{zeng2019end}, and add differentiable tracking in-the-loop \cite{liang2019multi}. While these works are great at dealing with uncertainties at the sensor level and mitigating failures downstream from object detection uncertainty by learning joint features, they do not focus on the output parameterization, and all of them produce uni-modal predictions. Uni-modal predictions can result in unsafe behaviors if, for example, the predicted intention is not accurate (e.g., a pedestrian  crossing vs waiting) or the predictions lie inbetween two modes (e.g. at branching roads).
In this work, we extend this framework  to predict multi-modal behaviors that are aligned with human prior knowledge.

The motion planning algorithms in an SDV need to take into account all possibilities to make safe decisions. Thus, motion forecasting models that can characterize complex multi-modal distributions are a must, and efficient sampling is desired such that motion planning can timely find the plan with the lowest expected cost.
Approaches that directly output the parameters of the marginal distribution at each timestep with a closed-form likelihood such as a sequence of Gaussians over time \cite{alahi2016social, casas2019spatially} meet the efficient sampling requirement, but can suffer from low expressivity. \cite{cui2019multimodal} proposed a more stable way to train mixtures of future trajectories than directly optimizing the likelihood, by only training the Gaussian likelihood of the closest mode to the ground-truth trajectory and applying cross-entropy to the mode probabilities. \cite{chai2019multipath} followed up on this idea by anchoring the predictions to clusters produced offline by running k-means on the training set. These two models have a good tradeoff between sampling efficiency and expressivity. Predicting discrete occupancy maps into the future \cite{jain2019} has very good expressivity and naturally captures multi-modality, but is very memory intensive and requires adaptations in order to use traditional motion planners designed for trajectory representations.

Autoregressive models \cite{rhinehart2018r2p2, 2019arXiv190501296R, tang2019multiple} require sequential sampling and thus are not amenable to real-time inference, which is necessary in safety critical applications such as self-driving. 
Furthermore, there is a mismatch between training and inference: the distributions at training are conditioned on ground truth, but during inference they are conditioned on model samples. This makes the model less robust to noise, particularly in the joint detection and prediction setting where perfect detection is not possible.
Finally, modeling interactions between the traffic participants  \cite{alahi2016social, deo2018convolutional, casas2019spatially, 2019arXiv190501296R, tang2019multiple} has been shown to help reduce the prediction uncertainty, and generate more socially consistent predictions.

There has been incredible progress in learning multimodal futures with a likelihood objective without leveraging prior knowledge or known structure, thanks to the availability of large and diverse driving datasets. 
However, datasets collected from real-world driving can only cover one possible mode out of many possibilities of how the future might have unrolled. As shown in  \cite{rhinehart2018r2p2}, this partial description of the underlying distribution along with a Maximum Likelihood Estimation (MLE) objective results in high entropy distributions that favors mode-covering over mode-seeking behaviors \cite{bishop2006pattern}.
Empirically, although the learned distribution recovers the ground truth future, it also generates highly implausible samples (e.g., out-of-map predictions).
In this paper we show that coverage at the expense of precision severely harms motion planning. 

A solution to avoid mode-covering predictions is to directly leverage prior knowledge to characterize the true distribution. In particular, the actor's motion history, road geometry, traffic rules, and nearby traffic participants all constrain the space of plausible futures. Past approaches have (i) devised neural networks with inductive priors at the model level \cite{chang2019argoverse}, and (ii) use a map-based reconstruction loss as an auxiliary task (e.g. recovering the road mask shape \cite{bansal2018chauffeurnet} to encourage prediction to fall on drivable areas, which they show helps with generalization to novel driving scenarios). However, their approach is limited since the policy is non probabilistic, and the loss is applied to only the maximum a posteriori (MAP) estimate. 
In this paper, we propose a more general and powerful approach by leveraging the REINFORCE gradient estimator to incorporate any prior knowledge (including non-differentiable functions) over probabilistic motion forecasts, still allowing the model to recover non-compliant behavior.

\section{Exploiting Prior Knowledge in Driving}
\label{sec:prior}

\begin{figure}[t]
    \centering
    \includegraphics[width=\linewidth]{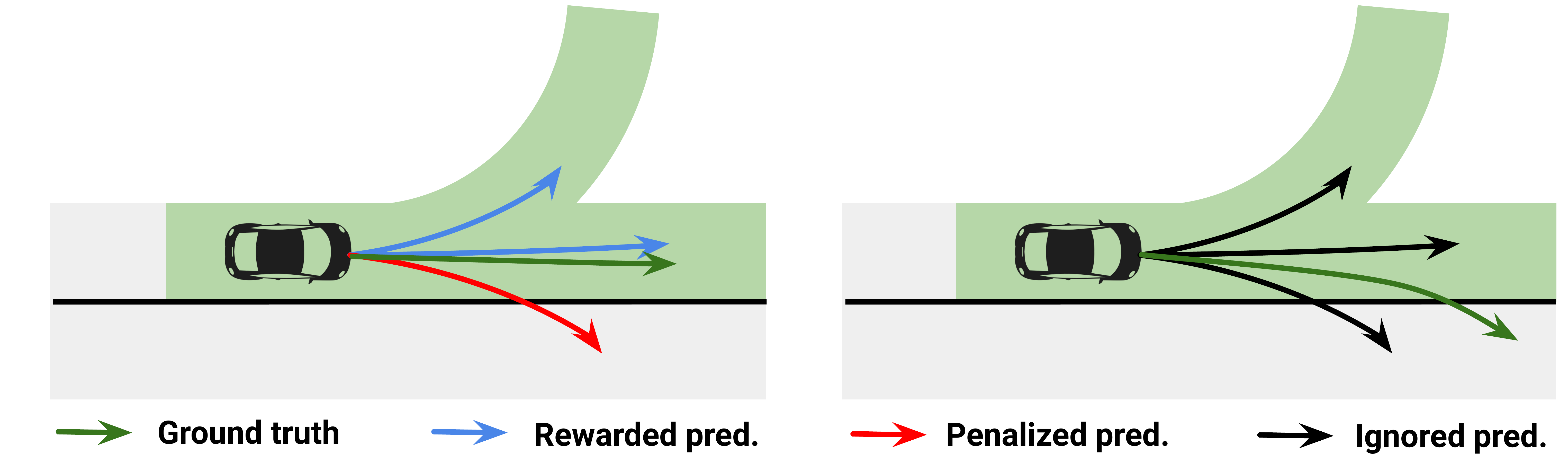}
    \caption{Reachable lanes (green surface) loss}
    \label{fig:reachable_loss}
    \vspace{-0.3cm}
\end{figure}

In this section, we describe a novel framework to incorporate prior knowledge explicitly into probabilistic motion forecasts. Importantly, our approach still permits the model to predict non-compliant behavior that does not follow traffic rules in the rare event that this occurs. 
Our method is general and can be applied to any model that can generate actor trajectory samples $y_n$ and evaluate their marginal likelihood $p(y_n|x_n)$ efficiently. Since for the remainder of the paper we only refer to per-actor marginal likelihoods, we simplify the notation and refer to an actor's trajectory as $y$, and its local context as $x$ (i.e. the detected bounding box and local LiDAR/map features).
We defer the explanation of the particular state-of-the-art perception and prediction model we use to Section~\ref{pnp}, as this is not the main contribution of our work. %

\begin{figure}[t]
    \centering
    \includegraphics[width=\linewidth]{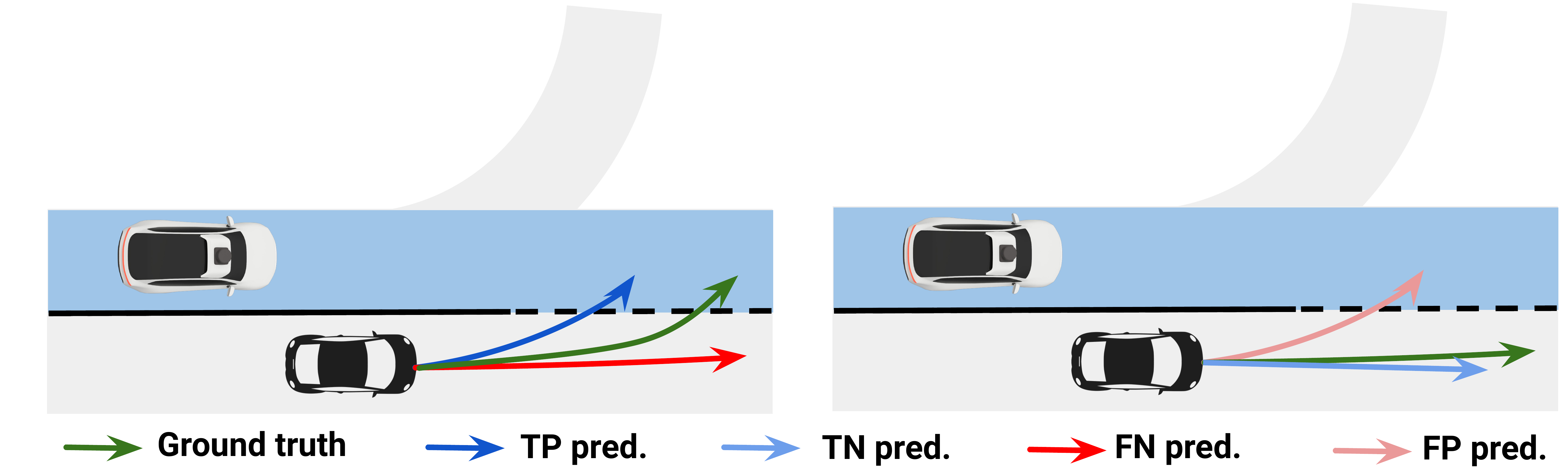}
    \caption{SDV Route (blue surface) loss}
    \label{fig:route_loss}
    \vspace{-0.3cm}
\end{figure}

Given a traffic scene, humans have rich prior knowledge over how the traffic participants might behave. In this paper, we propose to directly use this prior knowledge as supervision when learning an actor's distribution over future trajectories.
Towards this goal, we encode the prior knowledge as a deterministic {\it reward function}  $r(y, x)$. 
We then define the prior knowledge objective as the negative expected reward over samples from the future trajectory distribution. Note that applying the loss directly to the point estimate of the means is not sufficient, since our goal is to learn an accurate characterization of the full distribution for safe motion planning. 
The goal is then to learn a stochastic policy (parameterized by $\theta$) that maximizes the expected reward:
\begin{align*}
\mathcal{L}_\text{prior}(x;\theta) &= \mathop{\mathbb{E}_{y \sim p_\theta(y|x)}}\left[-r\left(y, x)\right)\right] \\
				   &= \int -p_\theta\left(y | x\right)r\left(y , x\right)dy
\end{align*}

Most priors are non-differentiable and cannot be easily relaxed (e.g., motion forecast following the traffic rules or not) Thus, we leverage policy gradient algorithms, which do not assume differentiability of the reward function $r$ and allow direct optimization without making any approximations. In particular, we use the popular REINFORCE algorithm \cite{williams1992simple}, which  only requires the policy to be differentiable, provide efficient sampling, and allow likelihood evaluation. In this case, the gradient of the expected loss can be computed as:
\begin{align*}
\nabla \mathcal{L}_\text{prior}(x;\theta) = 
\mathop{\mathbb{E}_{y \sim p_\theta\left(y|x\right)}}
\left[-\nabla \log p_\theta\left(y | x\right)
r\left(y, x\right)\right]
\end{align*}
The expectation can then be approximated by drawing samples from the predicted distribution as follows:
$$
\nabla \mathcal{L}_\text{prior}(x;\theta) \approx \frac{1}{S}\sum_i^S -\nabla \log p_\theta\left(y^i | x\right) r(y^i, x)
$$
where $y^i$ is the $i$th trajectory sample, over $S$ samples. 
Although this Monte Carlo estimation is unbiased, it typically has high variance. Our experiments  show that this does not pose a problem when using a policy that has an efficient sampling mechanism, since we can draw a large number of samples.

In practice, our  reward explicitly incorporates the prior knowledge that drivers generally follow the reachable lanes defined by lane markers, traffic signs and traffic lights.
Furthermore, we leverage our knowledge about the self driving vehicle's planned route to place emphasis on the most relevant actors. Intuitively, missing the prediction of an actor coming in conflict with the SDV route (false negatives) or predicting that an actor will cross in front of the SDV when in reality it stops (false positives) are of greater importance than accurately characterizing the behavior of an actor 50 meters behind the SDV. 
We express the final reward as a simple linear combination of rewards
$$r(y, x) 
= \sum_t^T r_{\text{reach}}(y_t, x) + r_{\text{route}}(y_t, x)$$ 
Next, we explain both terms in detail.

\subsection{Reachable Lanes}
Human driving behavior is highly structured: in the majority of scenarios, drivers will follow the road topology and traffic rules. To leverage this informative prior, but not overly penalize non-compliant behavior, we define a flexible traffic-rule informed loss that is conditioned on \emph{ground-truth} behavior. 
To this end, we leverage a lane-graph representation where the nodes encode lane segments and the edges represent relationships between lane segments such as adjacency, predecessor, and successor (taking into account direction of traffic flow).  This allow us to define the reachable lanes loss per timestep as:
\begin{equation}
    r_{\text{reach}}\left(y_t, x\right) =
    \begin{cases}
    r_d, & \text{if } y_t \in \text{reach}(x) \land y_t^{gt} \in \text{reach}(x)\\
    -r_d & \text{if } y_t \notin \text{reach}(x) \land y_t^{gt} \in \text{reach}(x)\\
    0, & \text{otherwise}
    \end{cases}\\
\end{equation}
where $\text{reach}(x)$ represent the lanes that are reachable from the detected vehicle bounding box by obeying the traffic rules. 
Note that to be robust to noise in the lane graph and avoid penalizing non-compliant behaviors, we only apply the loss if the ground truth waypoint $y_t^{gt}$ falls within the set of reachable lanes as well. 
This loss is summarized in Fig.~\ref{fig:reachable_loss}. 
To define this set of reachable lanes, we capture lane divider infractions as well as traffic light violations on the lane-graph.

\begin{table*}[ht]
	\centering
	\small
	\begin{threeparttable}
        \begin{tabularx}{\textwidth}{
                        >{\centering\arraybackslash}c |  %
                        >{\centering\arraybackslash}X 
                        >{\centering\arraybackslash}X 
                        >{\centering\arraybackslash}X |
                        >{\centering\arraybackslash}X 
                        >{\centering\arraybackslash}X 
                        >{\centering\arraybackslash}X |
                        >{\centering\arraybackslash}X 
                        >{\centering\arraybackslash}X 
                        >{\centering\arraybackslash}X }
		    \toprule
                Model & 
                \multicolumn{3}{c|}{Final Lane Error (\%)} &
                \multicolumn{3}{c|}{meanADE (m)} &
                \multicolumn{3}{c}{minADE (m)} \\
                {} & 
                Straight & Left & Right
                & Straight & Left & Right
                & Straight & Left & Right \\
            \midrule
                \textsc{SpAGNN} \cite{casas2019spatially}   & 18.79 & 51.27 & 51.52 & 3.36 & 5.70 & 6.22 & 0.68 & 1.63 & 1.68 \\
                \textsc{MultiPath} \cite{chai2019multipath} & 12.76 & 49.27 & 39.72 & 2.40 & 5.09 & 5.08 & 0.58 & 1.59 & 1.30 \\
                \textsc{R2P2-MA} \cite{2019arXiv190501296R} & 6.36 & 42.26 & 28.58 & 2.57 & 4.74 & 5.09 & 0.75 & 1.85 & 1.63 \\
                \midrule
                \textsc{SpAGNN+}                            & 10.07 & 47.92 & 39.60 & 2.35 & 4.53 & 4.89 & \textbf{0.53} & \textbf{1.39} & \textbf{1.26} \\
                \midrule
                \textsc{Ours}   & \textbf{6.28} & \textbf{39.13} & \textbf{28.07} & \textbf{2.17} & \textbf{4.16} & \textbf{4.57} & 0.54 & 1.60 & 1.51 \\
	        \bottomrule
		\end{tabularx}
	\end{threeparttable}
	\caption{[\ourdataset] Motion forecasting}
	\label{table:tor4d_main}
\end{table*}
\begin{table}[t]
	\centering
	\small
	\begin{threeparttable}
        \begin{tabularx}{\linewidth}{
                        >{\centering\arraybackslash}c |  %
                        >{\centering\arraybackslash}X 
                        >{\centering\arraybackslash}X 
                        >{\centering\arraybackslash}X |
                        >{\centering\arraybackslash}X |
                        >{\centering\arraybackslash}X }
		    \toprule
                Model & 
                \multicolumn{3}{c|}{Final Lane Error (\%)} &
                \multicolumn{1}{c|}{meanADE} &
                \multicolumn{1}{c}{minADE} \\
                & 
                Straight & Left & Right &
                {(m)} & 
                {(m)} \\
            \midrule
                \textsc{SpAGNN+} & 15.24 & 24.24 & 28.48 & 1.72 & \textbf{0.48} \\
                \midrule
                \textsc{Ours} & \textbf{10.69} & \textbf{17.44} & \textbf{20.31} & \textbf{1.65} & 0.50 \\
	        \bottomrule
        \end{tabularx}
    \end{threeparttable}

	\caption{[\nuscenes] Motion Forecasting}
    \label{table:nuscenes_main}
    \vspace{-0.3cm}
\end{table}

\subsubsection{Lane infraction} Lane dividers limit the set of legal high-level actions a vehicle can take in the road. For instance, lane changing over a solid line or taking over another vehicle by crossing a yellow double solid line into opposite traffic are not allowed. We incorporate this prior by removing the edges corresponding to illegal maneuvers from the lane-graph. Encoding this prior helps the model predict less entropic distributions.

\subsubsection{Traffic light violation} Many interactions occur at intersections, some of which are safety critical. Thus it is important to have accurate actor predictions at intersections, particularly differentiating stopping and going behaviors. To this end, we leverage the traffic control states (i.e., green, red, yellow) to remove edges connecting lane segments that are currently governed by a red traffic light.

Once we have processed the lane-graph by applying the aforementioned rules, we perform lane association to match each vehicle bounding box to a lane (or set of lanes when the vehicle overlaps with multiple ones, for example during a lane change). Subsequently, we run a depth first search starting from the current lane segment, obtaining a set of reachable lanes.

\subsection{SDV Route}
It is more important to precisely characterize the motion of vehicles that might interact with the SDV \cite{refaat2019agent}, rather than other traffic participants that do not influence the SDV behavior.
We can approximate the area of interest with the SDV's route (i.e. high-level command), which is defined as the union of all lane segments that the SDV can travel on to reach a preset goal, given the lane-graph. More concretely, the horizon is set to be equal to the prediction horizon (5s), and the target lane is generated by a given high level route planner (out of the scope of this paper). 
This gives a safe approximation over its future possible locations.

We now define positive trajectories as those with at least one waypoint falling within the SDV route, and negative otherwise. We would like our trajectory predictions to achieve high precision and high recall under this definition, taking into account if the ground-truth trajectory intersects the route (positive) or not (negative). 

More concretely, we define the route loss as:
\begin{equation}
    r_{\text{route}}\left(y_t, x\right) =
    \begin{cases}
      r_{tp}, & \text{if } y_t \in \text{route} \land y_t^{gt} \in \text{route}\\
      r_{fp}, & \text{if } y_t \in \text{route} \land y_t^{gt} \notin \text{route} \\
      r_{tn}, & \text{if } y_t \notin \text{route} \land y_t^{gt} \notin \text{route} \\
      r_{fn}, & \text{if } y_t \notin \text{route} \land y_t^{gt} \in \text{route} \\ 
    \end{cases}
\end{equation}
where we have different rewards for true positive, false positive, true negative and false negative waypoint predictions, since there is high imbalance in the data and they have different impact on the safety of our motion planner. This loss is illustrated in Fig.~\ref{fig:route_loss}.

\section{Perception and Prediction Model} \label{pnp}

So far we have presented our general framework, but we have not specified the particular perception and motion forecasting model we apply our prior knowledge loss to. We exploit a combination of the backbone feature extraction, object detection network, and graph propagation from \textsc{SpAGNN} \cite{casas2019spatially}, together with the mixture of Gaussians output parameterization of \textsc{MTP} \cite{cui2019multimodal}. This perception and prediction model 
takes a voxelized LiDAR point cloud and a raster map as input, extracts scene features using a backbone CNN, and applies Rotated Region of Interest Align (RRoI) \cite{ma2018arbitrary} to extract per-actor features. After that, we define a fully-connected graph where the nodes correspond to traffic participants, and perform a series of graph-propagations to refine each actor's representation by aggregating features from their neighbors, as proposed by \cite{casas2019spatially}. Finally, an MLP header predicts the parameters of a multimodal distribution over future trajectories by using a mixture of Gaussians with full covariance for each actor: 
\begin{align*}
p(y|x) = \sum_{k} p(a^k |x) \prod_{t=1}^{T}p(y_t | \mu_{t}^k (x), \Sigma_{t}^k (x))
\end{align*}

Our method can be trained end-to-end using back-propagation and stochastic gradient descent. In particular, we minimize a multi-objective loss containing a classification and regression terms for object detection, 
a symmetric motion forecasting loss agnostic to the map or SDV, as well as the prior informed non-differentiable loss we described in Section~\ref{sec:prior}. The loss of each actor is a weighted sum of the multiple objectives.
\begin{gather*}
\mathcal{L} = \alpha \cdot \mathcal{L}_{\text{det}} + \beta \cdot \mathcal{L}_{\text{symmetric}} + \gamma \cdot \mathcal{L}_{\text{prior}}
\end{gather*}

For the classification branch ($\mathcal{L}_{\text{cla}}$) of the detection header (background vs vehicle), we employ a binary cross entropy loss with hard negative mining. In particular, we select all positive examples from the ground-truth and 3 times as many negative examples from the rest of spatial locations. For box fitting ($\mathcal{L}_{\text{reg}}$), we apply a smooth L1 loss to each of the 5 parameters $(x, y, w, h, \phi)$ of the bounding boxes anchored to a positive example .
\begin{gather*}
\mathcal{L}_{\text{det}} = \mathcal{L}_{\text{cla}} + \lambda \cdot \mathcal{L}_{\text{reg}}
\end{gather*}

Instead of directly optimizing the likelihood of the mixture model, we follow \cite{cui2019multimodal} in heuristically matching the closest mode to the ground-truth and only taking the negative log likelihood of that mode, while training the mode scores with a cross-entropy loss. This has been shown empirically \cite{chai2019multipath} to be a more stable training objective than optimizing the mixture likelihood directly as in \cite{bishop1994mixture}. Thus we define
\begin{align*}
    \mathcal{L}_{\text{symmetric}} = - \sum_{k} \text{1}(k=\hat{k})[\log p(a^k | x) + \sum^T_t \log p(y^k_t | x)]
\end{align*}
where $\hat{k} = \argmin_{k} \text{dist}(y^k, y^{gt})$ is the mode whose mean is closest to the ground truth trajectory in euclidean distance.

Since the gaussian waypoints in this model are independent across time, we use a heuristic trajectory sampler at inference to draw smooth samples from this model (see Appendix \ref{appendix} for details). 
Note that during training we do not need this sampler since all the losses are formulated at the waypoint level. 
However, it is important to have temporally consistent samples for measuring the impact of our predictions in the downstream task of motion planning, since we approximate the heading into the future by finite differences between waypoints of the sample trajectories.

\begin{table*}[t]
	\centering
	\small
	\begin{threeparttable}
        \begin{tabularx}{\linewidth}{
                        >{\centering\arraybackslash}c |   
                        >{\centering\arraybackslash}X |  
                        >{\centering\arraybackslash}X | 
                        >{\centering\arraybackslash}X | 
                        >{\centering\arraybackslash}X | 
                        >{\centering\arraybackslash}X  
                        }
		    \toprule
                Model & 
                \multicolumn{1}{c|}{Collision} & 
                \multicolumn{1}{c|}{L2 human} &
                \multicolumn{1}{c|}{Lat. acc.} &
                \multicolumn{1}{c|} {Jerk} & 
                \multicolumn{1}{c}{Progress} \\
                {} &
                \multicolumn{1}{c|}{(\% up to 5s)} & 
                \multicolumn{1}{c|}{(m @ 5s)} & 
                \multicolumn{1}{c|}{(m/$s^2$)} &
                \multicolumn{1}{c|}{(m/$s^3$)} &
                \multicolumn{1}{c}{(m @ 5s)} \\
            \midrule
                \textsc{SpAGNN} \cite{casas2019spatially}           & 4.19  & 5.98  & 2.94 & 2.90 & 32.37  \\
                \textsc{R2P2-MA} \cite{2019arXiv190501296R}         & 3.71  & 5.65  & 2.84 & 2.53 & \textbf{33.90} \\
                \textsc{MultiPath} \cite{chai2019multipath}         & 3.30  & 5.58  & 2.73 & 2.57 & 32.99 \\
            \midrule
                \textsc{SpAGNN+}                                    & 3.33  & 5.52  & 2.77 & 2.56  & 33.11  \\
            \midrule
                \textsc{Ours}   & \textbf{2.75}  & \textbf{5.43} & \textbf{2.67} & \textbf{2.47} & 33.09  \\
	        \bottomrule
		\end{tabularx}
	\end{threeparttable}
    \caption{[\ourdataset] System level performance}
    \vspace{-0.3cm}
	\label{table:planning}
\end{table*}

\vspace{-0.1cm}
\section{Experimental Evaluation}
\vspace{-0.1cm}

In this section, we first describe how we measure the motion forecasting ability of our method, namely how well it predicts the future behavior of all the traffic participants. 
We then introduce the comprehensive set of metrics that we use for evaluation in the downstream task of ego-motion planning. Despite being the final goal of the system, this task has generally been ignored in previous motion forecasting works.
Next, we discuss the state-of-the-art baselines we compare against, report extensive quantitative results on two challenging, real-world datasets: \ourdataset{} \cite{yang2018pixor} and \nuscenes{} \cite{caesar2019nuscenes}, and showcase the differences of adding prior knowledge from a qualitative standpoint.
Finally, we perform a thorough ablation study to justify our choices on how to incorporate prior knowledge. 
We defer our implementation details to Appendix \ref{appendix}.

\subsection{Metrics}

Motion forecasting results neglect the fact that actors are of different importance in the overall system, and that motion forecasting metrics are not necessarily aligned with system level metrics. A model with the best aggregate metrics in perception and motion forecasting may excel at the unimportant cases, and miss safety-critical cases resulting in unsafe driving. 
Since our method is a perception and motion forecasting system, we lead the discussion with motion forecasting metrics for clarity, as well as to gain intuitions on what are the differences with the baselines.
However, what we really care about is how well we drive, and thus provide an extensive analysis of system level metrics to conclude this section.

\subsubsection{Motion Forecasting}
We follow previous works \cite{casas2018intentnet, yang2018pixor, casas2019spatially} in joint perception and prediction and perform IoU-based matching between object detections and ground-truth bounding boxes, ignoring fully occluded vehicles without any LiDAR point. 

In order to measure the performance of the motion forecasting system we use sample quality measures, following \cite{2019arXiv190501296R}. In particular, we use (i) the final lane error (trajectory waypoint inside vs. outside the reachable lanes at 5 seconds into the future) to measure map understanding, (ii) minimum average displacement error (minADE) to show the recall of our motion forecasts at different time horizons, and (iii) mean average displacement error (meanADE), which gives us an idea of the precision of our predictions since unrealistic samples severely harm this metric. Furthermore, we benchmark the performance in a diverse set of driving behaviors, by breaking down all metrics by the ground-truth high-level action of the vehicle: going straight, left turning and right turning. We omit stationary and near stationary vehicles, since all models do well in detecting and predicting the future states of those. We use 50 samples for all evaluations. 

Note that, in contrast to \cite{2019arXiv190501296R}, we emphasize the need of metrics that show the precision of the predictions, as opposed to using only recall-oriented metrics such as minADE, for 2 reasons. First, recall is easily achievable at the expense of precision by simply predicting very fan-out distributions. Second, precision in the motion forecasts is critical for safe motion planning, as we show in Sec.~\ref{plan_exp}.

\subsubsection{Ego-Motion Planning}
To evaluate how our approach impacts the full system, we use the  learnable motion planner proposed in \cite{sadat2019jointly}.
We feed the planner with 50 trajectory samples for each vehicle, as a Monte Carlo approximation of the marginal distribution. Thus, we assign equal weight to each sample to avoid overweighing the high likelihood region of the distribution, since the samples already come from our model. 
This way we can keep the motion planner as proposed. 
Because in this paper we do not consider perception and prediction of pedestrian and bicyclists, we feed the ground-truth trajectories of these traffic participants to the motion planner, for all experiments.

We focus on the safety-related metric of collision rate ($\%$ of time the SDV plan collides with any other traffic participant in the ground-truth, for a future horizon of 5 seconds). We also provide results on comfort-related metrics such as lateral acceleration and jerk, to reveal any potential tradeoff. Finally, we also include the progress of the SDV to show that the methods are not just trivially avoiding collisions and uncomfortable situations by staying still. Note that these metrics are computed in open-loop, by unrolling the motion plan for the duration of the prediction horizon.

\begin{table*}[ht]
    \vspace{-0.3cm}
    \begin{tabular}{@{}c@{\hspace{.1em}}c@{\hspace{.5em}}c@{\hspace{.5em}}c@{\hspace{.5em}}c}
        \textbf{} & \textbf{Example 1} & \textbf{Example 2} & \textbf{Example 3} & \textbf{Example 4} \\

        \rotatebox[origin=c]{90}{\textbf{\textsc{SpAGNN+}}} &
        \raisebox{-0.5\height}{\includegraphics[trim={3cm 1cm 3cm 1cm},clip, width=0.24\textwidth]{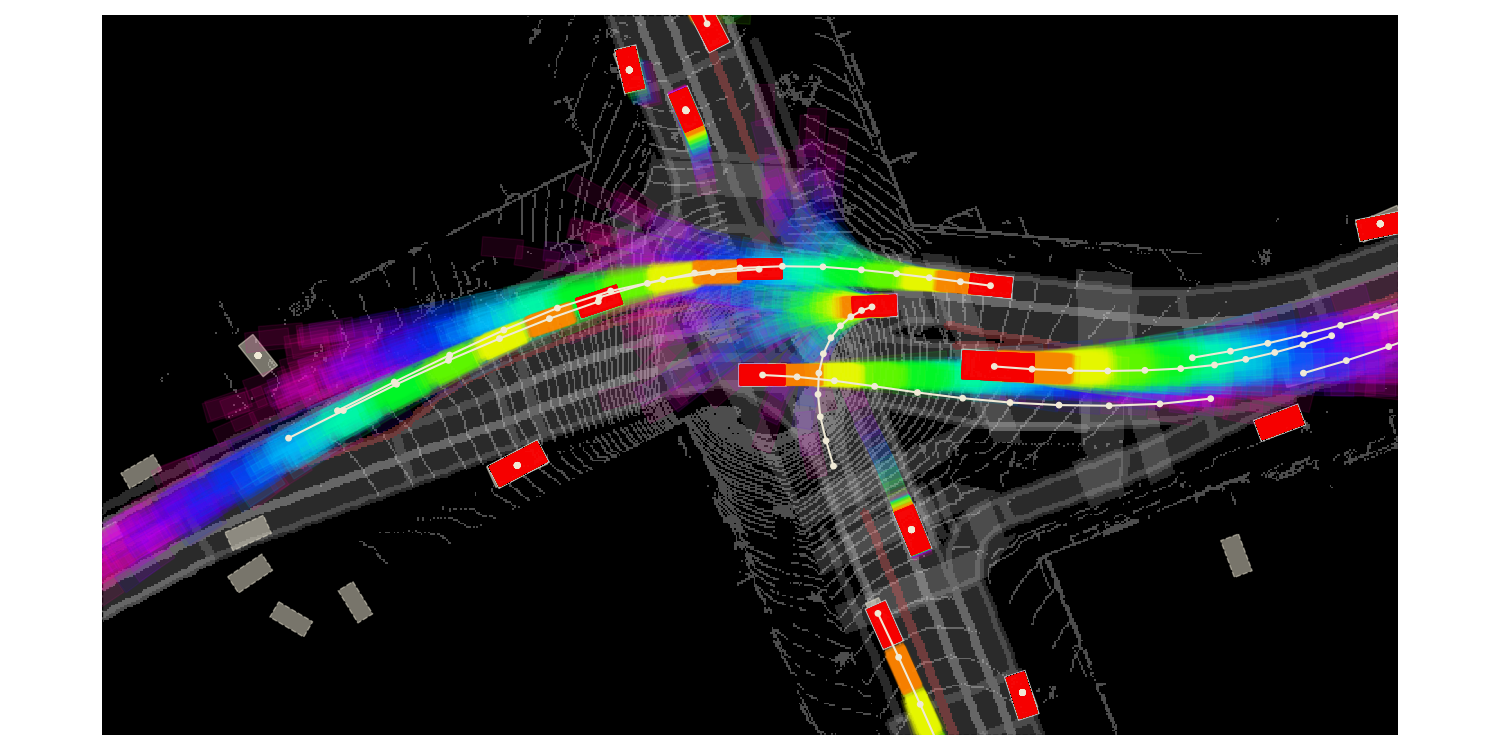}} & 
        \raisebox{-0.5\height}{\includegraphics[trim={3cm 1cm 3cm 1cm},clip, width=0.24\textwidth]{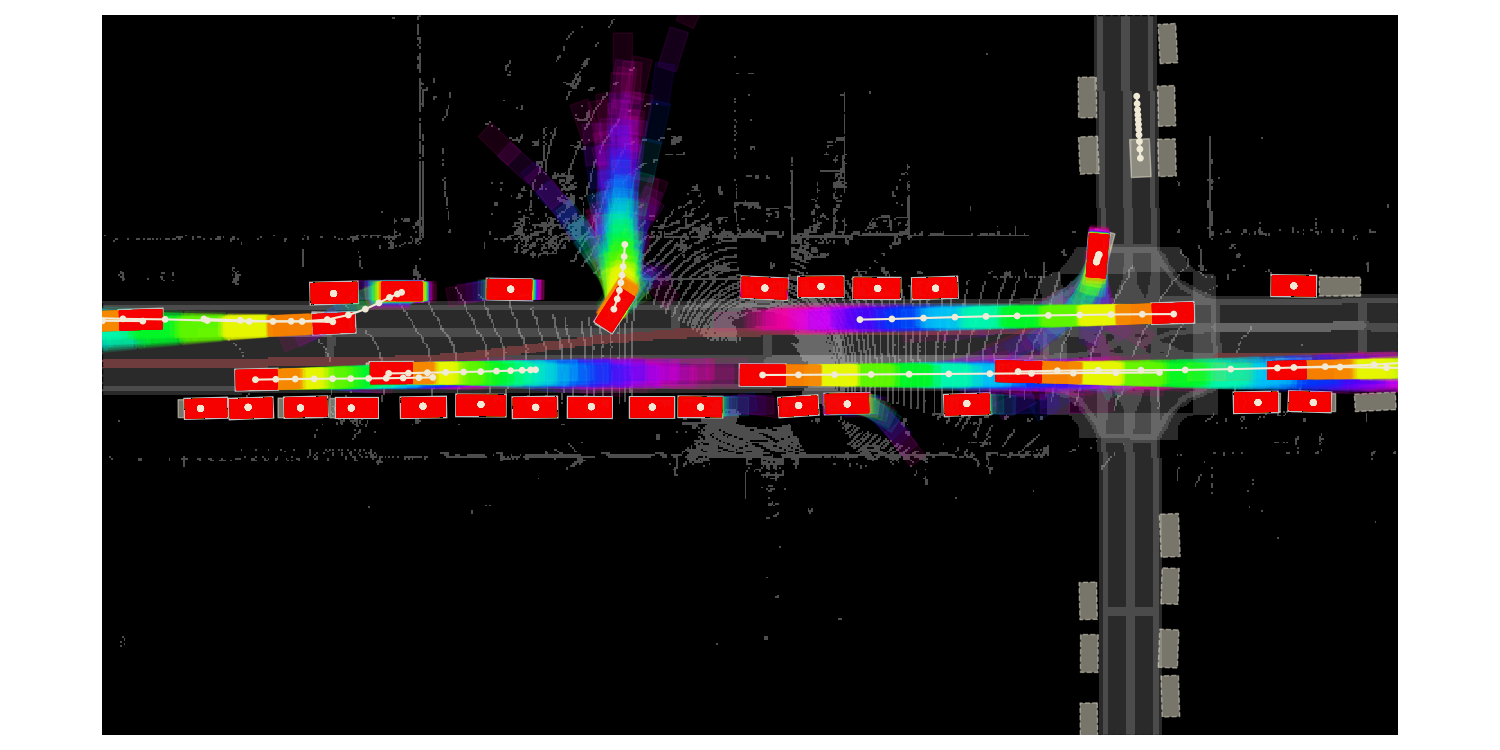}} &
        \raisebox{-0.5\height}{\includegraphics[trim={3cm 1cm 3cm 1cm},clip, width=0.24\textwidth]{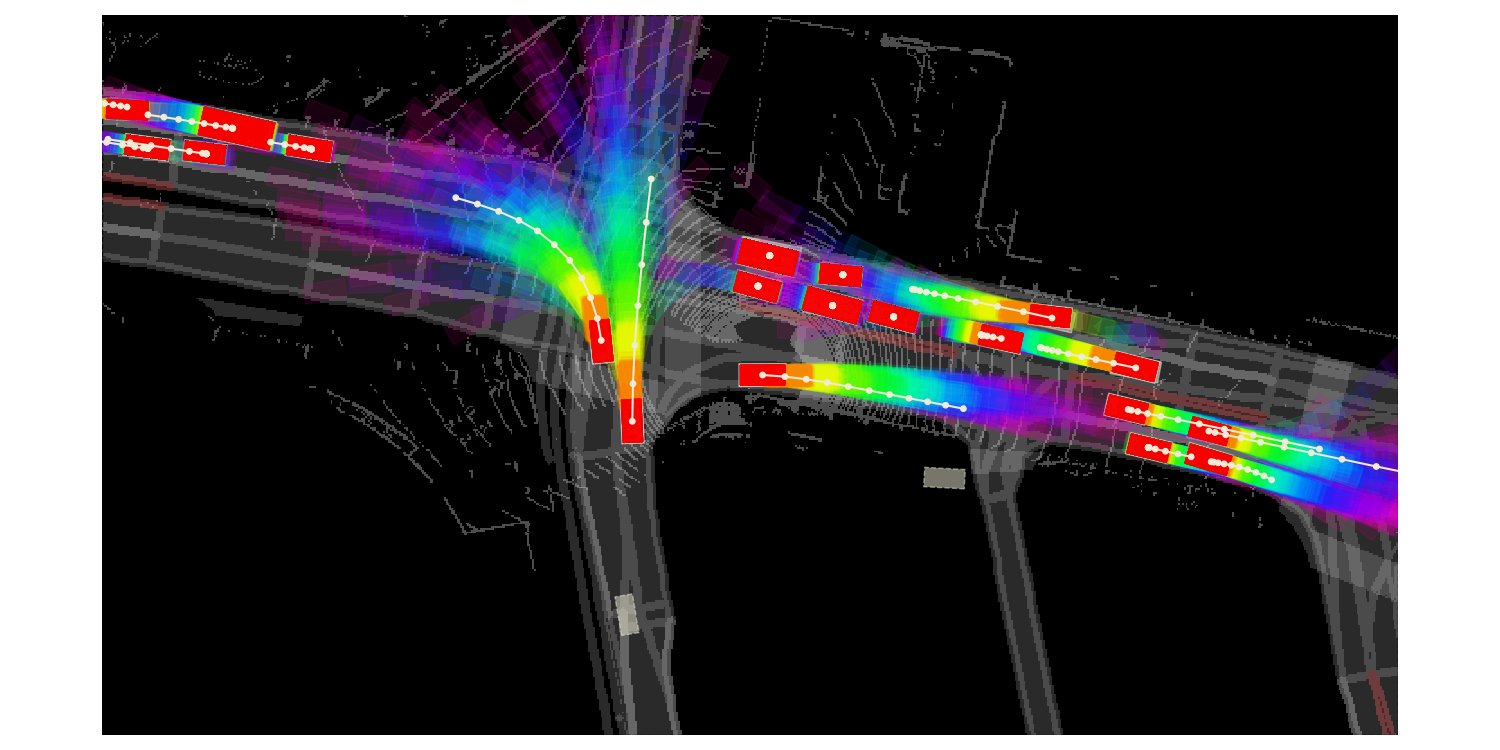}} &
        \raisebox{-0.5\height}{\includegraphics[trim={3cm 1cm 3cm 1cm},clip, width=0.24\textwidth]{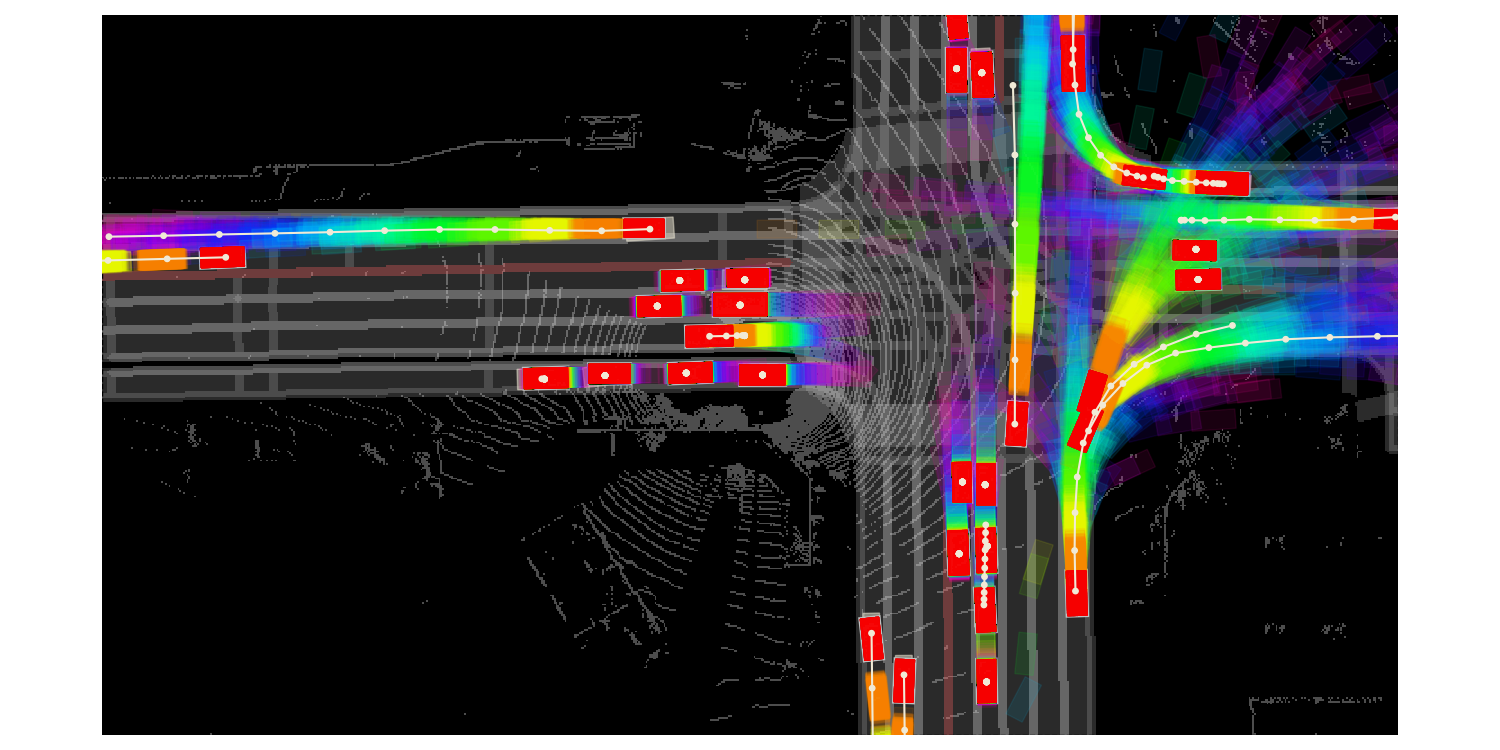}} \vspace{.5em} \\

        \rotatebox[origin=c]{90}{\textbf{\textsc{Ours}}} &
        \raisebox{-0.5\height}{\includegraphics[trim={3cm 1cm 3cm 1cm},clip, width=0.24\textwidth]{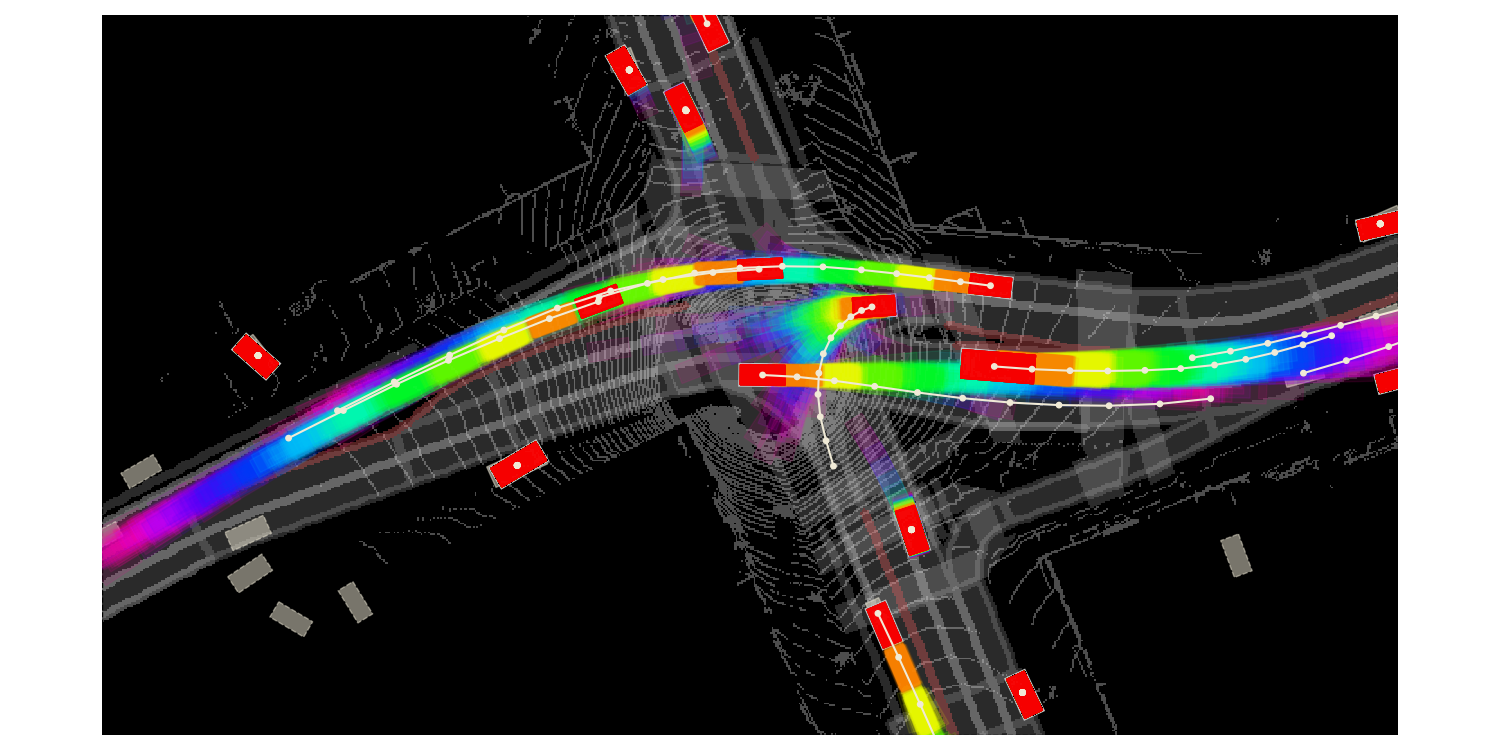}} &
        \raisebox{-0.5\height}{\includegraphics[trim={3cm 1cm 3cm 1cm},clip, width=0.24\textwidth]{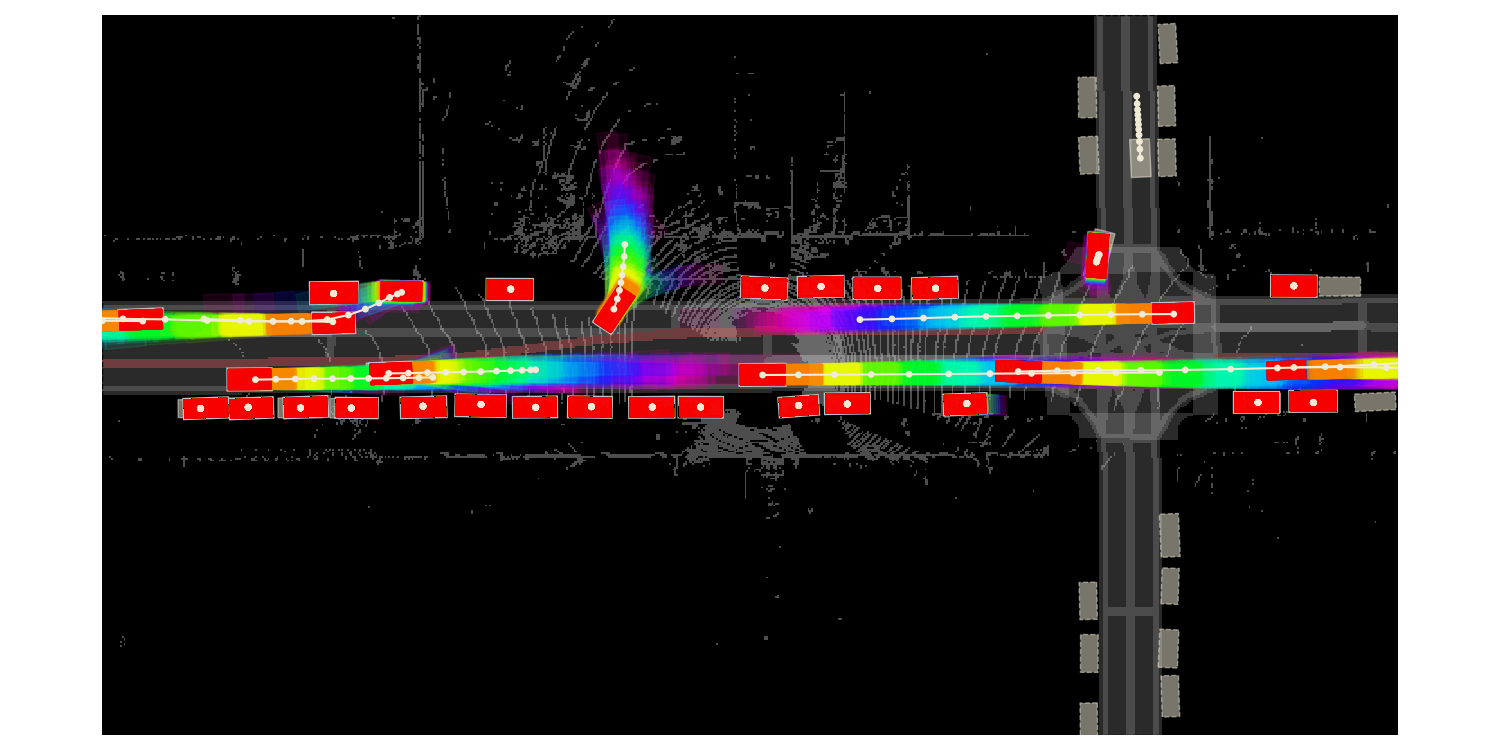}} &
        \raisebox{-0.5\height}{\includegraphics[trim={3cm 1cm 3cm 1cm},clip, width=0.24\textwidth]{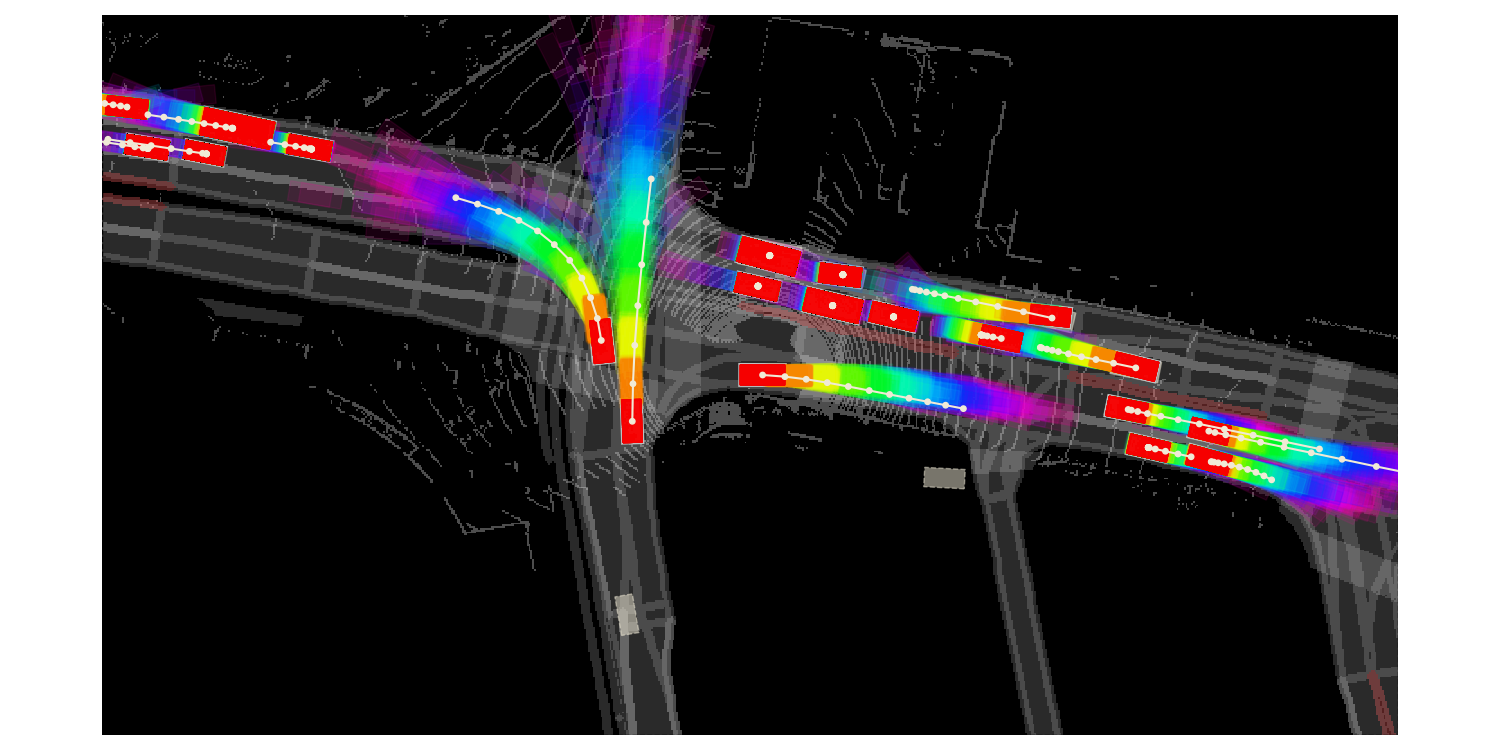}} &
        \raisebox{-0.5\height}{\includegraphics[trim={3cm 1cm 3cm 1cm},clip, width=0.24\textwidth]{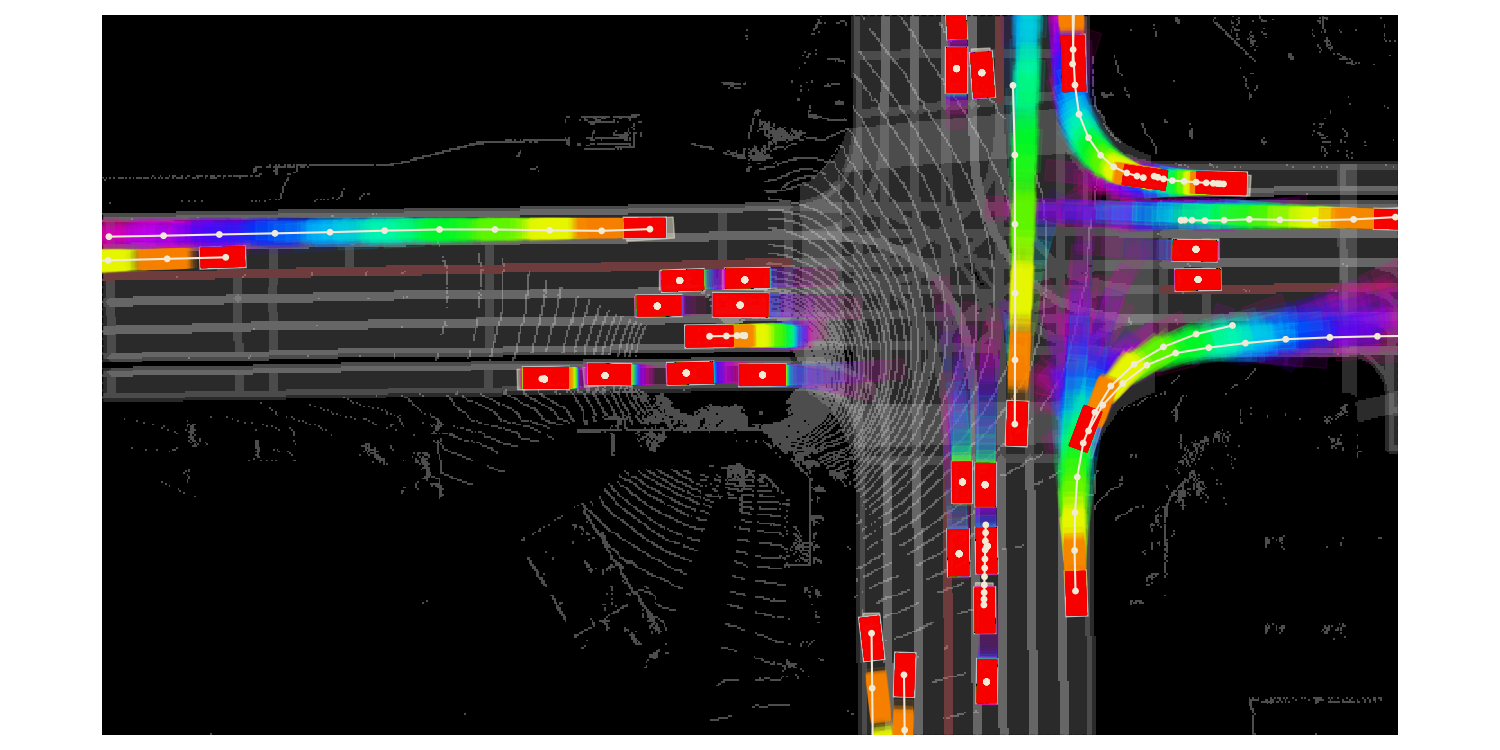}} \\
    \end{tabular}
    \caption{Motion forecasting visualizations of 50 smamples. Time in the motion forecasts is encoded in the rainbow colormap ranging from red (0s) to pink (5s). Ground-truth is shown by the white trajectory waypoints.}
    \label{fig:pred}
\end{table*}

\begin{table*}[t]
    \vspace{-0.2cm}
    \begin{tabular}{@{}c@{\hspace{.1em}}c@{\hspace{.5em}}c@{\hspace{.5em}}c@{\hspace{.5em}}c}
        \textbf{} &
        \textbf{t = 0 s} &
        \textbf{t = 1.5 s} &
        \textbf{t = 3.5 s} &
        \textbf{t = 5 s} \\

        \rotatebox[origin=c]{90}{\textbf{\textsc{SpAGNN+}}} &
        \raisebox{-0.5\height}{\includegraphics[trim={3cm 1cm 3cm 1cm},clip, width=0.24\textwidth]{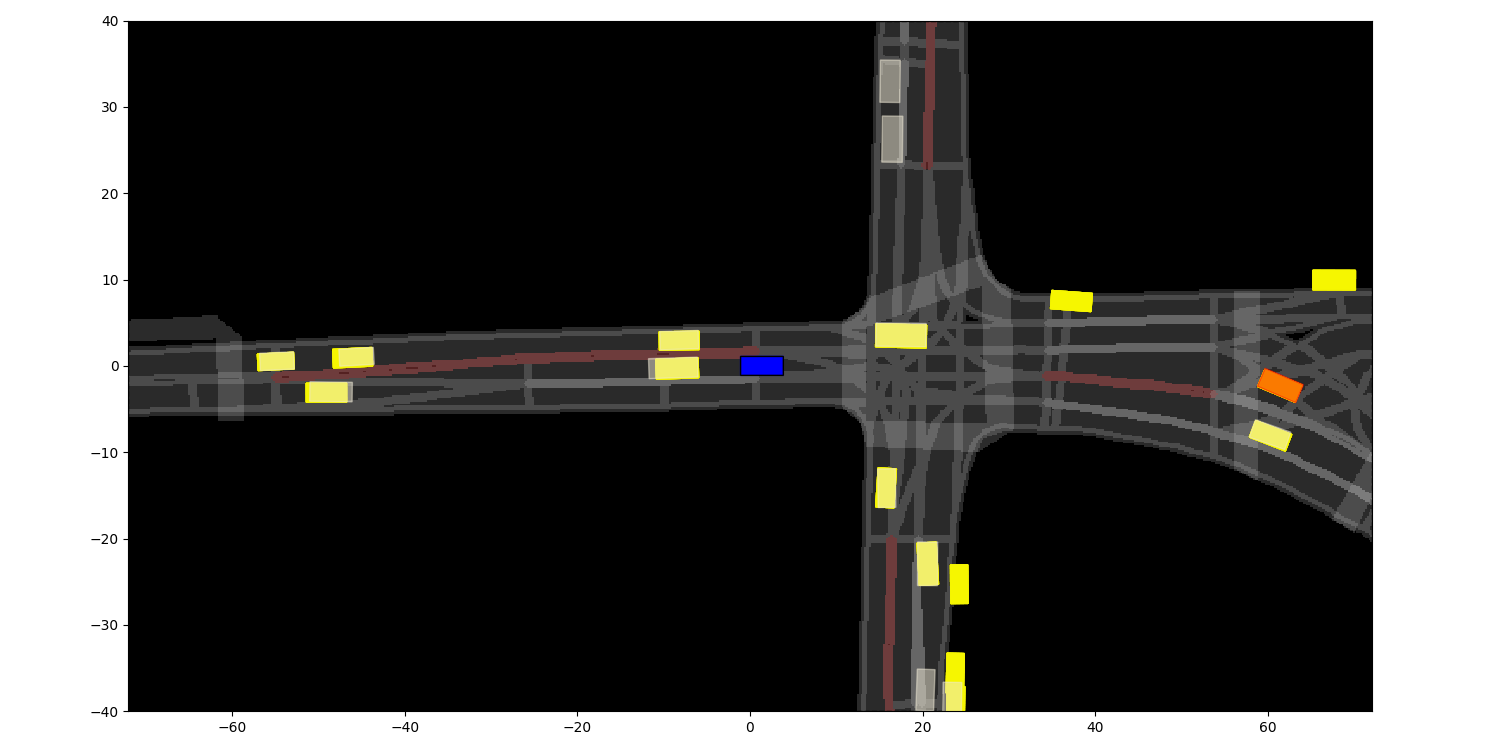}} &
        \raisebox{-0.5\height}{\includegraphics[trim={3cm 1cm 3cm 1cm},clip, width=0.24\textwidth]{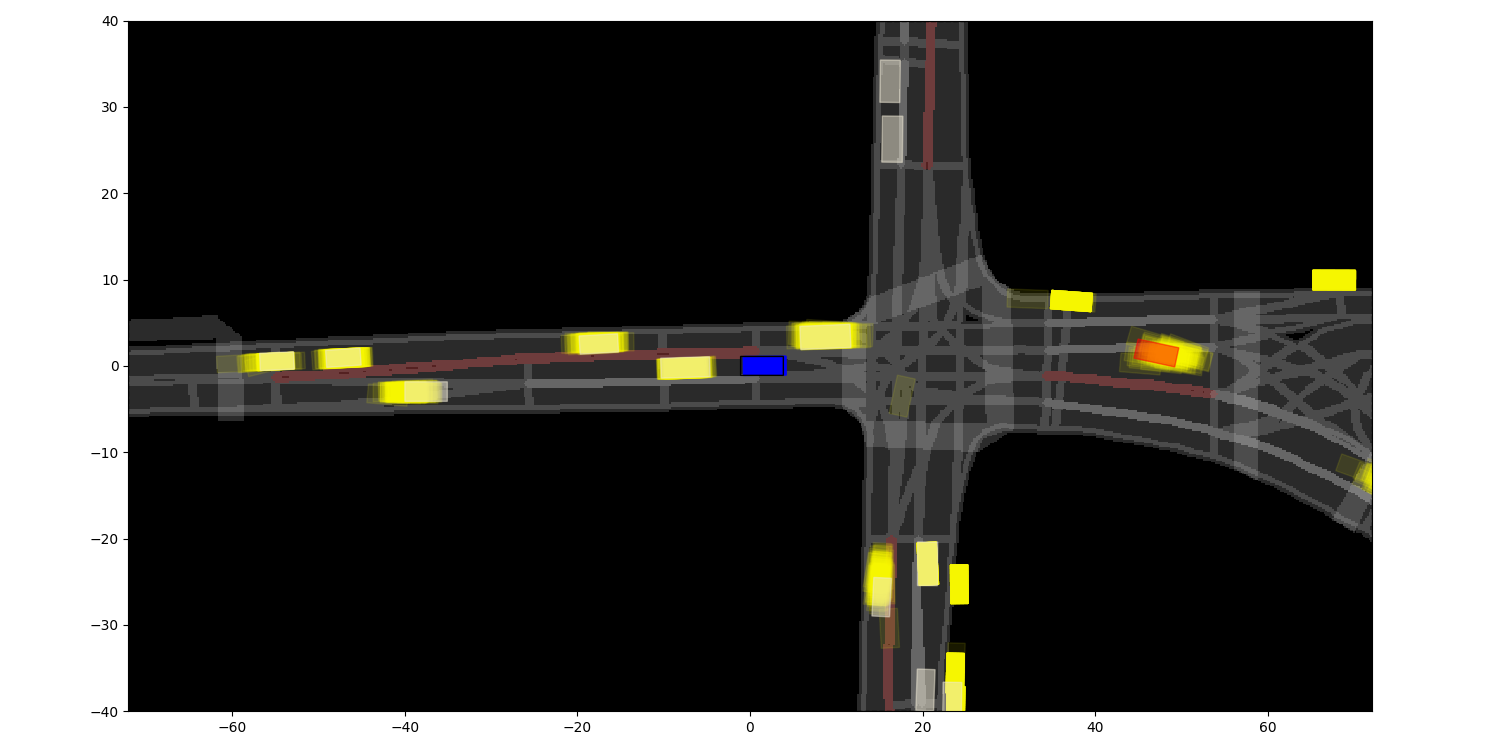}} &
        \raisebox{-0.5\height}{\includegraphics[trim={3cm 1cm 3cm 1cm},clip, width=0.24\textwidth]{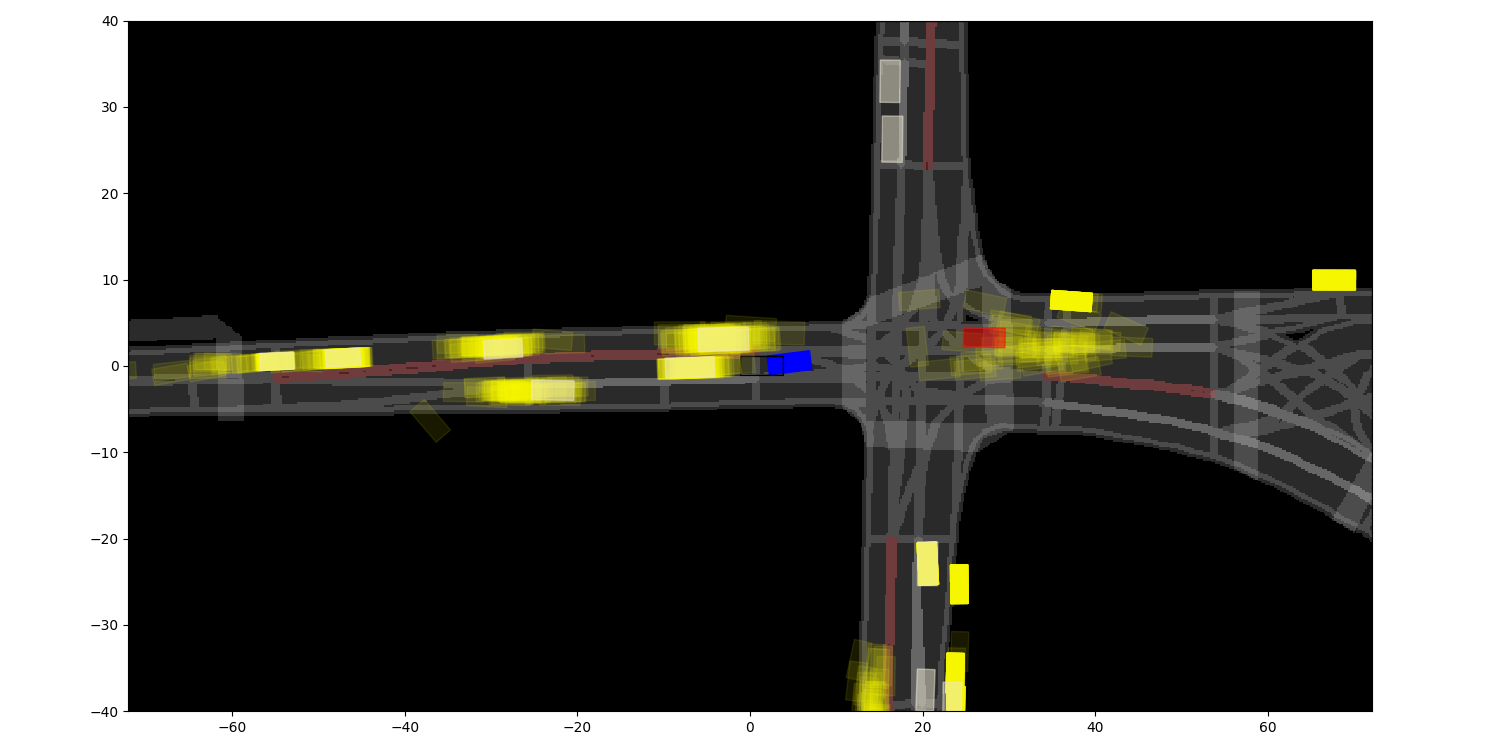}} &
        \raisebox{-0.5\height}{\includegraphics[trim={3cm 1cm 3cm 1cm},clip, width=0.24\textwidth]{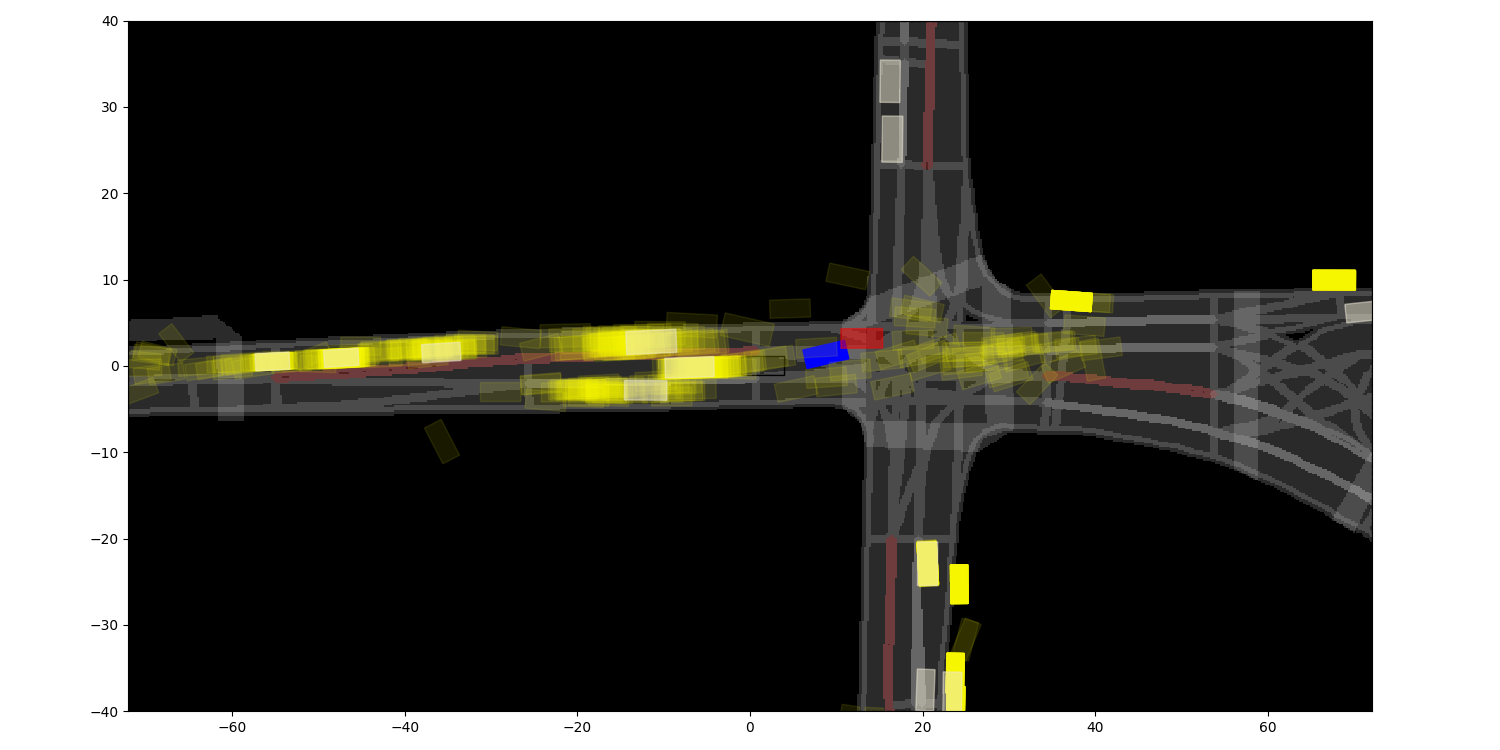}} \vspace{.5em} \\
        
        \rotatebox[origin=c]{90}{\textbf{\textsc{Ours}}} &
        \raisebox{-0.5\height}{\includegraphics[trim={3cm 1cm 3cm 1cm},clip, width=0.24\textwidth]{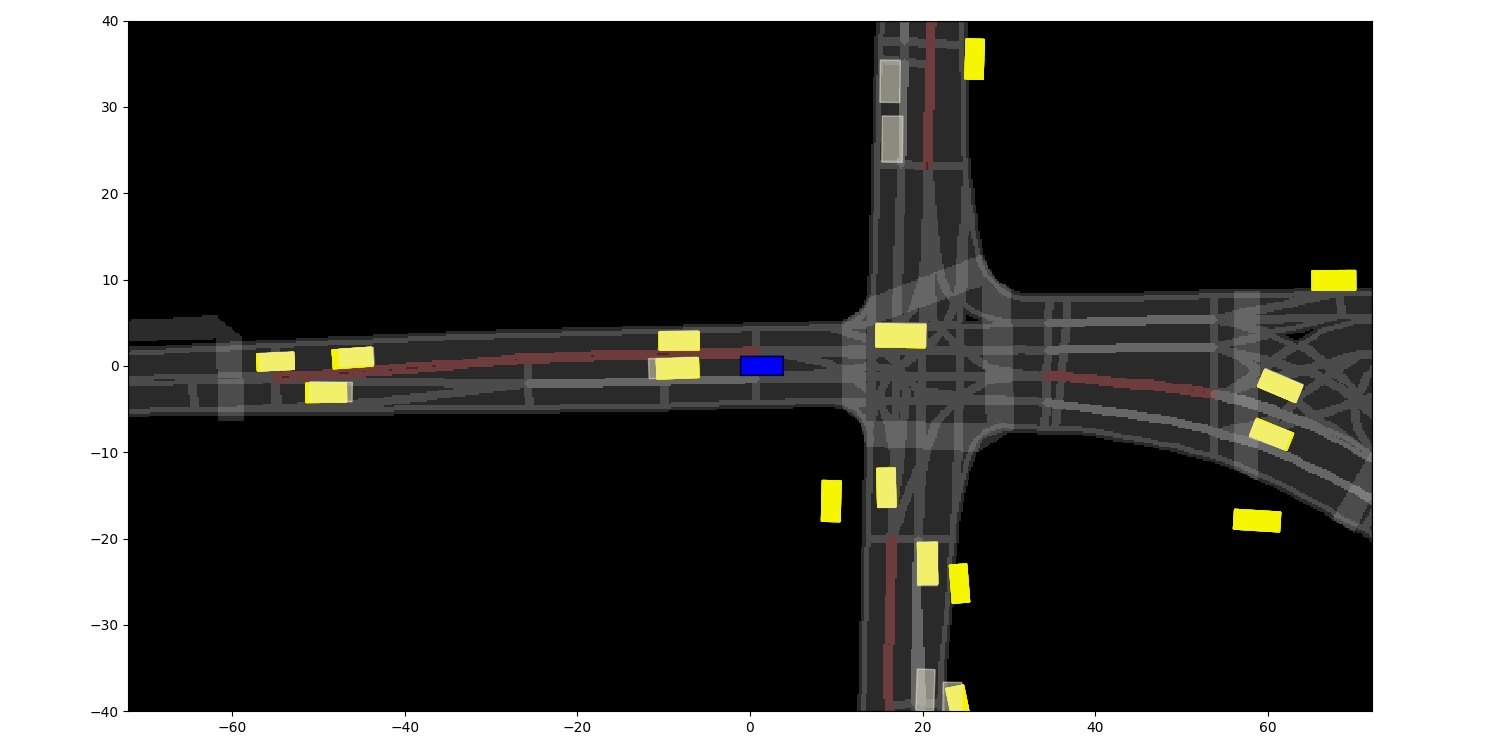}} &
        \raisebox{-0.5\height}{\includegraphics[trim={3cm 1cm 3cm 1cm},clip, width=0.24\textwidth]{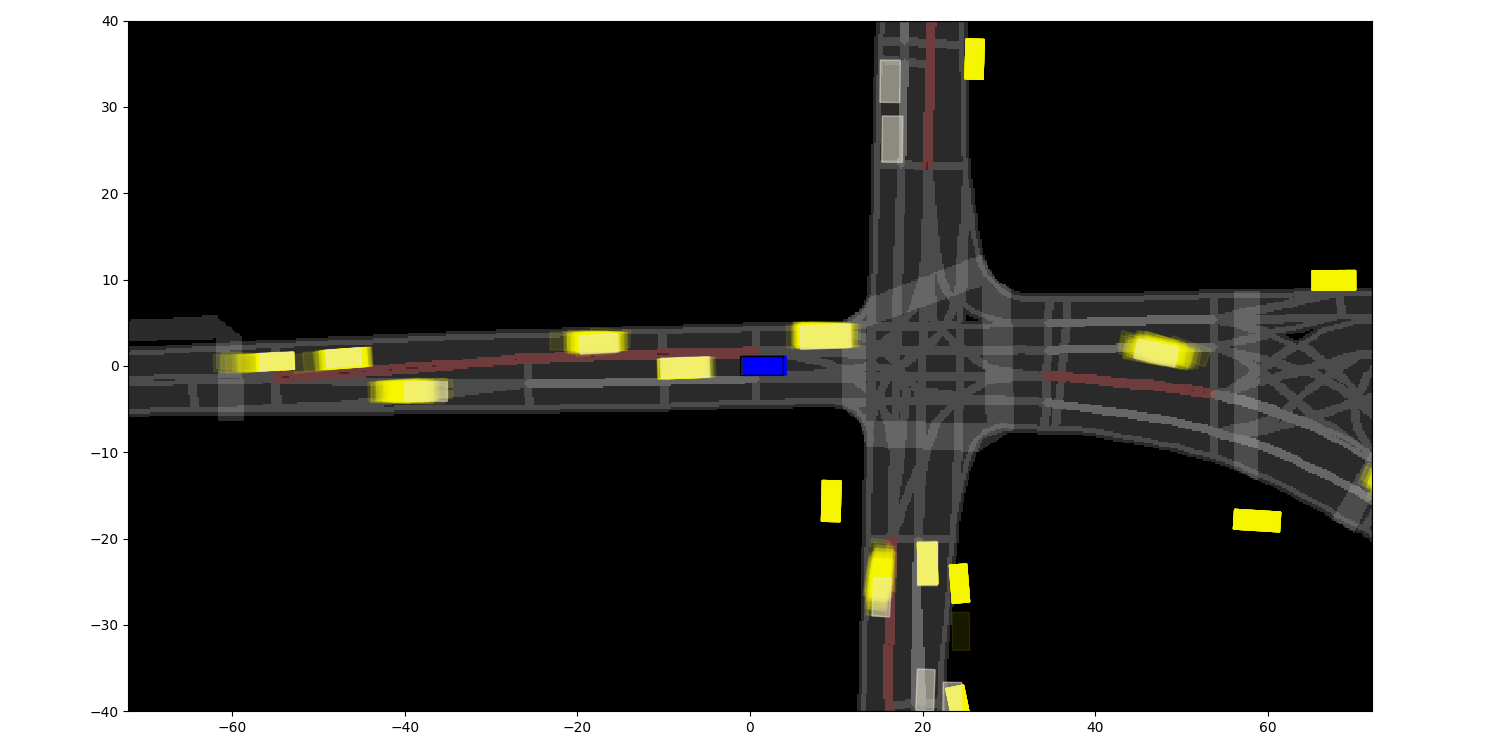}} &
        \raisebox{-0.5\height}{\includegraphics[trim={3cm 1cm 3cm 1cm},clip, width=0.24\textwidth]{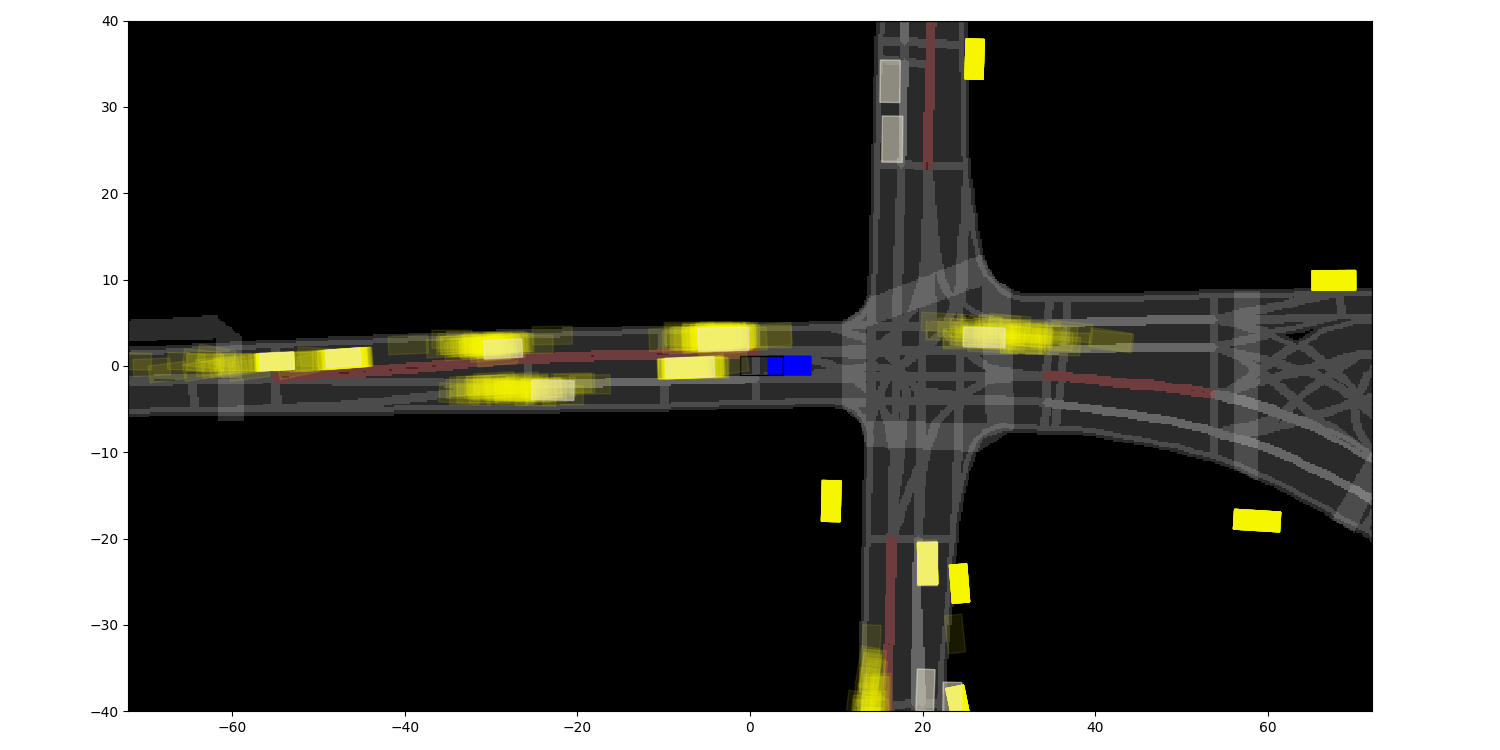}} &
        \raisebox{-0.5\height}{\includegraphics[trim={3cm 1cm 3cm 1cm},clip, width=0.24\textwidth]{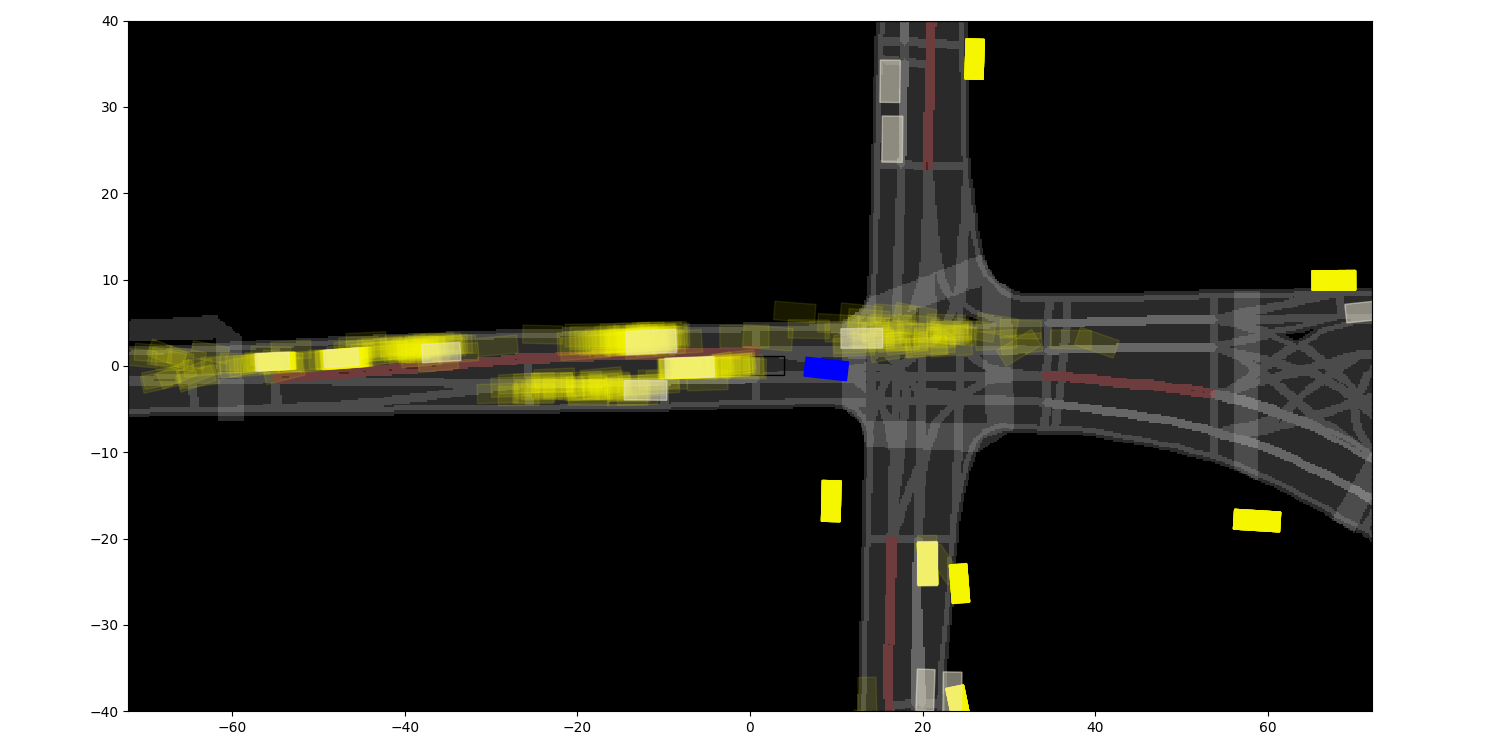}} \\
    \end{tabular}
    \caption{An ego-vehicle harmful event is avoided with more precise motion forecasts. The predicted bounding boxes into the future for other traffic participants are shown in yellow if not colliding, and in red if colliding with the SDV plan (shown in blue). Ground-truth future trajectories are shown in white, and ground-truth SDV trajectory as an empty black box.}
    \label{figure:planning_comparison}
    \vspace{-0.3cm}
\end{table*}

\subsection{Comparison Against State of the Art} \label{plan_exp}

We compare our approach to three previously proposed state-of-the-art motion forecasting approaches: \textsc{SpAGNN} \cite{casas2019spatially}, \textsc{R2P2-MA} \cite{2019arXiv190501296R}, and \textsc{MultiPath}  \cite{chai2019multipath}. 
We also consider the model outlined in Section \ref{pnp} but without applying our proposed prior knowledge loss ($\mathcal{L}_{\text{prior}}$) as a baseline, which we call \textsc{SpAGNN+}.
We implemented all models in the context of joint perception and prediction, and employ the same detection backbone in the baselines as in our approach for a fair comparison. We provide more details about the adaptations of \textsc{R2P2-MA} \cite{2019arXiv190501296R}, and \textsc{MultiPath}  \cite{chai2019multipath} to the joint perception and prediction setting in Appendix \ref{appendix}.

\noindent
\textbf{Motion Forecasting:}
Table \ref{table:tor4d_main} shows motion forecasting results in \ourdataset{}, operating at a 90 $\%$ common recall point across all models for fair comparison. We first show that adding the mixture of gaussians as output parameterization to SpAGNN improves all sample quality metrics (SpAGNN vs. SpAGNN+). This makes sense since unimodal distributions cannot capture multi-modal behaviors such as breaking vs. accelerating, or turning right vs. going straight. 
Incorporating prior knowledge via our proposed method delivers much better map understanding and precision, as shown by the final lane error and meanADE. 
While the distance between the ground-truth and the closest sample (minADE) suffers a minor regression when incorporating prior knowledge, we have shown that this does not impact the downstream task of motion planning in Section~\ref{plan_exp}. Indeed, it is a metric that does not provide a description of the full distribution and overly favors those with higher entropy.

We show qualitative results for 4 scenarios in \ourdataset{} with diverse road topologies in Fig.~\ref{fig:pred}. We can clearly see how adding prior knowledge makes the distributions less entropic while preserving multi-modality, and substantially improves the map understanding of our predictions. We highlight that despite incorporating prior knowledge about the fact that vehicles tend to follow the map, we can see in \textit{Example 2} how our model can predict vehicles going out of the map (in this case an unmapped driveway).

In Table~\ref{table:nuscenes_main}, we validate our method on \nuscenes{} to show that it is robust to variations in the high definition map specification. We use the same evaluation setup, with the exception that we operate our object detector at a 60$\%$ common recall point. As in our dataset, our approach shows improvements in precision and map understanding metrics.

\noindent
\textbf{Ego-Motion Planning:}
In Table~\ref{table:planning}, we show that our method yields much safer motion plans, indicated by a 17\% collision reduction over the strongest baseline in this metric. Remarkably, the increase in safety is not at the expense of comfort, where our method achieves marginally less jerk and lateral acceleration than other approaches. Note that while progress in general is desired, it cannot be at the expense of safety and comfort. We notice that the ego-motion plans make similar progress across models, but our approach produces the closest trajectories to the ground truth executed by an expert human driver (lowest L2 distance at 5 seconds into the future), while yielding much fewer collisions. We observe that despite the popularity of the minADE metric across previous works in motion forecasting, the model that achieves the lowest minADE in Table \ref{table:tor4d_main} does not yield the safest or most comfortable ego-motion plans.

Fig.~\ref{figure:planning_comparison} showcases an example that illustrates a repeated behavior throughout the dataset: the baseline method produces much more fan-out distributions that cause the ego-vehicle to verge into dangerous situations. In this particular example, there is a vehicle in the oncoming lane for which the baseline predicts with significant probability that will drive into the SDV's lane, making the motion planner to drive into opposite traffic and finally causing a collision. In contrast, our model predicts a more precise and plausible distribution over possible futures where the relevant vehicle is predicted to follow its original lane.

\subsection{Ablations}

To show that our design choices to incorporate  prior knowledge are sound, we explore different approaches:

\begin{table*}[ht]
    \vspace{-0.5cm}
	\centering
	\small
	\begin{threeparttable}
        \begin{tabularx}{\textwidth}{
                        >{\centering\arraybackslash}c |  %
                        >{\centering\arraybackslash}c |  %
                        >{\centering\arraybackslash}X 
                        >{\centering\arraybackslash}X 
                        >{\centering\arraybackslash}X |
                        >{\centering\arraybackslash}X 
                        >{\centering\arraybackslash}X 
                        >{\centering\arraybackslash}X |
                        >{\centering\arraybackslash}X 
                        >{\centering\arraybackslash}X 
                        >{\centering\arraybackslash}X }
		    \toprule
                Prior Repr. &
                Approach &
                \multicolumn{3}{c|}{Final Lane Error (\%)} &
                \multicolumn{3}{c|}{meanADE (m)} &
                \multicolumn{3}{c}{minADE (m)} \\
                {} & {} &
                Straight & Left & Right
                & Straight & Left & Right
                & Straight & Left & Right \\
            \midrule
                - & -                  & 10.07 & 47.92 & 39.60 & 2.35 & 4.53 & 4.89 & 0.53 & 1.39 & 1.26 \\
                \midrule
                Reach. & Reconstruction        & 10.08 & 48.02 & 38.21 & 2.37 & 4.48 & 4.85 & \textbf{0.53} & \textbf{1.32} & \textbf{1.24} \\
                Centerline Dist & Mean  	   & 11.95 & 53.78 & 46.11 & 2.80 & 11.07 & 9.03 & 0.63 & 2.07 & 1.95 \\
				Reach. Dist & Mean 	   & 9.68 & 47.65 & 39.73 & 2.24 & 4.98 & 5.12 & 0.55 & 1.58 & 1.53 \\            
				Centerline Dist & Reparam		& 7.22 & 40.21 & 30.66 & 2.36 & 4.81 & 5.24 & 0.57 & 1.80 & 1.90 \\               
                Reach. Dist & Reparam 		& 6.84 & \textbf{37.75} & \textbf{27.55} & 2.29 & 4.46 & 4.91 & 0.56 & 1.68 & 1.86 \\                
                \midrule
                Reach. & REINFORCE		& 6.74 & 41.57 & 29.80 & 2.25 & 4.47 & 4.82 & 0.55 & 1.68 & 1.65 \\
                Reach. \& Route & REINFORCE    & \textbf{6.28} & 39.13 & 28.07 & \textbf{2.17} & \textbf{4.16} & \textbf{4.57} & 0.54 & 1.60 & 1.51 \\

	        \bottomrule
		\end{tabularx}
	\end{threeparttable}
	\caption{[\ourdataset] Motion forecasting ablations}
	\label{table:tor4d_ablation}
\end{table*}
\begin{table*}[ht]
	\centering
	\small
	\begin{threeparttable}
        \begin{tabularx}{\linewidth}{
                        >{\centering\arraybackslash}c |   
                        >{\centering\arraybackslash}c |   
                        >{\centering\arraybackslash}X |  
                        >{\centering\arraybackslash}X | 
                        >{\centering\arraybackslash}X | 
                        >{\centering\arraybackslash}X | 
                        >{\centering\arraybackslash}X  
                        }
		    \toprule
                Prior Repr. &
                Approach &
                \multicolumn{1}{c|}{Collision} & 
                \multicolumn{1}{c|}{L2 human} &
                \multicolumn{1}{c|}{Lat. acc.} &
                \multicolumn{1}{c|} {Jerk} & 
                \multicolumn{1}{c}{Progress} \\
                {} & 
                {} &
                \multicolumn{1}{c|}{(\% up to 5s)} & 
                \multicolumn{1}{c|}{(m @ 5s)} & 
                \multicolumn{1}{c|}{(m/$s^2$)} &
                \multicolumn{1}{c|}{(m/$s^3$)} &
                \multicolumn{1}{c}{(m @ 5s)} \\
            \midrule
                - & -           & 3.33  & 5.52 & 2.77 & 2.56  & \textbf{33.11}  \\
                \midrule
                Reach. & Reconstruction       & 3.58  & 5.54  & 2.80 & 2.56 & 33.10 \\
                Centerline Dist & Mean	    & 3.94  & 5.72  & 2.81 & 2.75 & 32.40\\  
                Reach. Dist & Mean         & 3.30  & 5.46  & 2.75 & 2.55 & 32.99 \\
                Centerline Dist & Reparam     & 3.61  & 5.52 & 2.70 & 2.56 & 33.00 \\ 
				Reach. Dist & Reparam         & 3.25  & 5.49 & 2.70 & 2.52 & 33.04 \\ 
                \midrule
                Reach. & REINFORCE      & 3.00  & 5.45 & 2.70 & 2.51 & 33.02  \\
                Reach. \& Route & REINFORCE    & \textbf{2.75}  & \textbf{5.43} & \textbf{2.67} & \textbf{2.47} & 33.09  \\
	        \bottomrule
		\end{tabularx}
	\end{threeparttable}
	\caption{[\ourdataset] System level performance ablations}
    \label{table:planning_ablation}
    \vspace{-0.3cm}
\end{table*}

\paragraph{Reachable Lanes Reconstruction} Similar to the road loss proposed in \cite{bansal2018chauffeurnet}, this approach uses an auxiliary convolutional head at the RRoI level that predicts the reachable lanes for each vehicle, represented as a spatial, binary mask. For this ablation, we replace our prior loss by a cross-entropy per pixel reconstruction loss, hoping to make the backbone features more map-aware.
\paragraph{Differentiable relaxations}
If the prior knowledge is a differentiable reward function $r$ over the support of the distribution, we can directly apply the loss to the model samples and backpropagate the gradients through the gaussian output distribution using the reparameterization trick \cite{kingma2013autoencoding}. 

\begin{figure}[t]
    \centering
    \includegraphics[width=\linewidth]{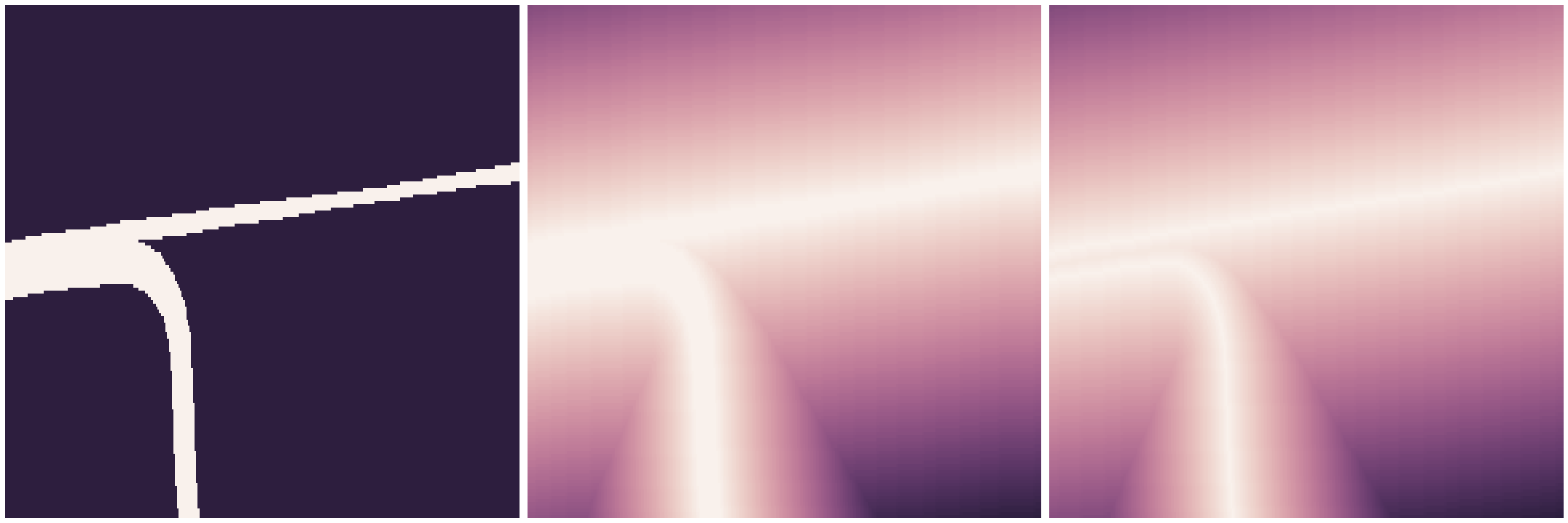}
    \caption{\textbf{Left}: Reachable Lanes, \textbf{Middle}: Distance to Reachable Lanes, \textbf{Right}: Distance to Centerline}
    \label{fig:relaxations}
    \vspace{-0.5cm}
\end{figure}

More concretely, we consider two differentiable relaxations of our non-differentiable reachable lanes loss as baselines, shown in Fig.~\ref{fig:relaxations}:
\begin{enumerate}
	\item Distance to centerline: we define the loss at each spatial location as the closest distance to a centerline of any lane in the set of reachable lanes.
	\item Distance to boundary: we define the loss at each spatial location as the closest distance to the boundary of one of the reachable lanes when outside the reachable lanes surface, and zero when inside.
\end{enumerate}

We test the aforementioned relaxed losses by applying them both to the mean of the distribution only, as well as to samples drawn from $p(y|x)$.

Next, we discuss the motion forecasting results and finally the impact on motion planning.

\noindent
\textbf{Motion Forecasting }
As shown in Table~\ref{table:tor4d_ablation}, the implicit feature learning approach inspired by \cite{bansal2018chauffeurnet} barely changes from the baseline. Applying the relaxed loss only to the mean of the distribution can have conflicting effects with the data likelihood term in the loss, since the latter optimizes the full distribution but the former only the mean. This yields to results that are worse than the base model over all metrics. Finally, applying the relaxed loss to the samples via reparameterization trick achieves the best lane metrics, but harms the meanADE. This is caused by the approximation of the reachable lanes loss to a continuous function when applying the distance transform. For instance, in curvy lanes or turns, the relaxed loss will push the predictions to the closest point in the reachable lanes, causing the trajectory to shorten (i.e. reduce the speed). 
Another drawback is that in branch topologies, the prediction will get pushed to the closest branch, which could differ from the ground-truth one.

\noindent
\textbf{Ego-Motion Planning }  
In Table~\ref{table:planning_ablation}, we show that not all approaches to incorporating prior knowledge improve the safety and comfort of ego-motion planning. 
In particular, the continuous relaxations of the reachable lanes loss do not reduce the number of collisions despite improving the final lane error as shown in the previous paragraph. We conjecture that there is a fine balance between map understanding, precision, and recall that is adequate for motion planning. The baselines sacrifice too much precision and recall in exchange of map understanding, most likely due to the approximations in the relaxation. Utilizing the REINFORCE gradient estimator to optimize the exact prior translates into much safer plans, particularly when incorporating the SDV route loss, showing that it is important to focus on the actors that can interact with the SDV.

\section{Conclusion and Future Work}

In this paper we have proposed a novel framework to explicitly incorporate prior knowledge into probabilistic motion forecasts, while allowing to predict non-compliant behavior when there is evidence. Our method is general, and can be applied to any model that can generate trajectory samples  and evaluate their marginal likelihood per actor. 
We have demonstrated the effectiveness of our approach in two challenging real-world datasets, significantly outperforming other state-of-the-art methods in both motion forecasting as well as in the downstream task of motion planning. Though we have chosen \textsc{SpAGNN} \cite{casas2019spatially} as the base model here, our method is general. We plan on integrating our approach with more base models, with a particular interest in joint distributions over actors where we can also apply our prior knowledge about interactions, such as the fact that vehicles do not generally collide.

\bibliographystyle{IEEEtran.bst}
\bibliography{egbib}

\appendix
\subsection{Implementation Details} \label{appendix}
\paragraph{Reachable lanes and Route} We represent the reachable lanes and the route as a spatial binary mask (25 meters by 70 meters at 4 meter/pixel resolution) in each vehicle's coordinate frame defined by its heading. This allows us to calculate the loss over samples efficiently. To evaluate our prior loss over a trajectory, we simply index into the raster at different waypoints.

\paragraph{Training} The hyperparameters we use for the loss function are: $\alpha=1.0$, $\beta=0.1$, $\gamma=0.1$, and $\lambda=0.5$. The rewards that we used are $r_d=1.0$, $r_{tp} = r_{fp} = r_{fn} = 1.0$, and $r_{tn}=0.1$. We use Adam optimizer with a base learning rate $5e-6$, which we increase linearly with the batch size. All models are trained from scratch.

\paragraph{Network Architecture} We use the architecture proposed in \cite{casas2019spatially}, where we modify the final regression layer to predict 16 modes. Each mode is composed by a mode score, and 11 Gaussian waypoints into the future (one every 0.5 seconds). Each waypoint is composed of 5 predictions: its centroid ($\mu_x, \mu_y$), its standard deviation ($\sigma_x, \sigma_y$), and its correlation $\rho$. The prediction coordinate frame is defined by the detected bounding box, where the x axis corresponds to the heading of the vehicle.

\paragraph{Heuristic Trajectory Sampler} \label{heuristic_sampler} We can heuristically sample smooth trajectories over time from a model that predicts independent gaussian waypoints over time, such as the one described in Section \ref{pnp}. More precisely, we extract temporally smooth trajectories by applying the re-parameterization trick for a bi-variate normal $y_t^s = \mu_t  + A_t \cdot \varepsilon^s$, where the model predicts a normal distribution $\mathcal{N}\left(y^s_t | \mu_t, \Sigma_t \right)$ per waypoint $t$, $(A_t)^T \cdot A_t = \Sigma_t$ is the cholesky decomposition of the covariance matrix, and $\varepsilon_s \sim \mathcal{N}(0, I)$ is the noise sampled from a standard bi-variate normal distribution. 
Note that the noise $\varepsilon^s$ is constant across time $t$ for a given sample $s$. Intuitively, having a constant noise across time steps allows us to sample waypoints whose relative location with respect to its predicted mean and covariance is constant across time (i.e. translated by the predicted mean and scaled by the predicted covariance per time). 

\paragraph{Reachable lanes in nuScenes} Recently, high definition maps were provided as an extension to the original \nuscenes{} dataset. Unfortunately, some of the road topology information is not provided.
In order to make use of the road topology to obtain the reachable lanes, we programmatically inferred such information by defining neighbors to be lane polygons that share vertices, share a direction of travel and are not separated by a road marker barring crossing. Successors are lanes or intersections that share terminating vertices and share traffic directions.

\paragraph{Baseline adaptations} \textsc{SpAGNN} \cite{casas2019spatially} did not require any adaptation because it was already proposed to work in the joint perception and prediction setting we operate. However, the formulation in \textsc{MultiPath} \cite{chai2019multipath} and \textsc{R2P2-MA} \cite{2019arXiv190501296R} assumes object detections and past tracked trajectories for every actor are given. Because our approach is tracking-free, and \textsc{SpAGNN} \cite{casas2019spatially} showed that the joint perception and prediction setting yields better results than using an off-the-shelf tracker to obtain past trajectories, we perform adaptations to these baselines. For \textsc{MultiPath} \cite{chai2019multipath}, the single difference is that our RRoI pooled features come from sensor data and raster map, while in their case the actor's past tracked trajectories are rasterized. For \textsc{R2P2-MA} \cite{2019arXiv190501296R}, we replace the RNN-based past trajectory encoder with the same per-actor feature extractor that is used in our method.

\end{document}